%% file: 0_rx-main.tex
\pdfoutput=1
\documentclass{article}

\usepackage[preprint]{neurips_2024}

\usepackage[utf8]{inputenc} 
\usepackage[T1]{fontenc}    
\usepackage{hyperref}       
\usepackage{url}            
\usepackage{booktabs}       
\usepackage{amsfonts}       
\usepackage{nicefrac}       
\usepackage{microtype}      
\usepackage{xcolor}         
\setcitestyle{numbers}

\usepackage{color}
\definecolor{light}{rgb}{0.4, 0.4, 0.4}
\def\light#1{{\color{light}#1}} 

\usepackage{subfig}
\usepackage{dirtytalk}
\usepackage{graphicx}
\usepackage{multirow}
\usepackage{amsmath}
\graphicspath{ {./figs/} }

\title{Toward Optimal Search and Retrieval for RAG}

\author{
    Alexandria Leto \\
    Department of Computer Science\\
    University of Colorado Boulder\\
    \texttt{alex.leto@colorado.edu}
    \And
    Cecilia Aguerrebere\\
    Intel Labs\\
    \texttt{cecilia.aguerrebere@intel.com}\\
    \And
    Ishwar Bhati\\
    Intel Labs\\
    \texttt{ishwar.s.bhati@intel.com}\\
    \And
    Ted Willke\\
    Intel Labs\\
    \texttt{ted.willke@intel.com}\\
    \AND
    Mariano Tepper\\
    Intel Labs\\
    \texttt{mariano.tepper@intel.com}\\
    \And
    Vy Ai Vo\\
    Intel Labs\\
    \texttt{vy.vo@intel.com}\\
}

\begin{document}

\maketitle

\begin{abstract}
Retrieval-augmented generation (RAG) is a promising method for addressing some of the memory-related challenges associated with Large Language Models (LLMs). Two separate systems form the RAG pipeline, the retriever and the reader, and the impact of each on downstream task performance is not well-understood. Here, we work towards the goal of understanding how retrievers can be optimized for RAG pipelines for common tasks such as Question Answering (QA). We conduct experiments focused on the relationship between retrieval and RAG performance on QA and attributed QA and unveil a number of insights useful to practitioners developing high-performance RAG pipelines. For example, lowering search accuracy has minor implications for RAG performance while potentially increasing retrieval speed and memory efficiency.

\end{abstract}
\input{1_rx-intro}

\input{2_rx-background}
\input{3_rx-methods}
\input{4_rx-results}
\input{5_rx-conclusion}

\begin{ack}
This work was completed as part of an internship at Intel Labs.

We thank Mihai Capot\u{a} for helpful feedback provided throughout the project and on this manuscript.
\end{ack}

\small
\bibliography{references}
\bibliographystyle{unsrt}


\appendix
\include{6_rx-appendix}

\end{document}

%% file: 1_rx-intro.tex
\section{Introduction}
\label{sec:intro}



Retrieval-augmented generation (RAG) \cite{Lewis_2020} is gaining popularity due to its ability to address some of the challenges with using Large Language Models (LLMs), including hallucinations \cite{huang2023surveyhallucinationlargelanguage} and outdated training data  \cite{Lewis_2020, wang2024knowledgeeditinglargelanguage}. RAG pipelines are made up of two disparate components: a retriever, which identifies documents relevant to a query from a given corpus, and a reader, which is typically an LLM prompted with a query, the text of the retrieved documents, and instructions to use this context to generate its response. However, it is unclear how a RAG pipeline's performance on downstream tasks can be attributed to each of these components \cite{Lewis_2020, huang2023surveyhallucinationlargelanguage}. 

In this work, we study the contributions of retrieval to downstream performance.\footnote{\url{https://www.github.com/intellabs/rag-retrieval-study}}. For this purpose, we evaluate pipelines with separately trained retriever and LLM components, as training retrieval-augmented models end-to-end is both more resource-intensive and obfuscates the contribution of the retriever itself. We aim to address questions that will enable practitioners to design retrieval systems tailored for use in RAG pipelines. For example, what are the weaknesses of the typical search and retrieval setup in RAG systems? Which search hyperparameters matter for RAG task performance?

We choose to evaluate RAG pipeline performance on both standard QA and attributed QA. In attributed QA, the model is instructed to cite supporting documents provided in the prompt when making factual claims \cite{Rashkin_Nikolaev_Lamm_Aroyo_Collins_Das_Petrov_Tomar_Turc_Reitter_2023, Gao_Yen_Yu_Chen_2023}. This task is interesting for its potential to boost the trustworthiness and verifiability of generated text \cite{liu-etal-2023-evaluating}.

We make four contributions: (1) We show how both QA performance and citation metrics vary with more retrieved documents, adding new data to a small literature on attributed QA with RAG. 
(2) We describe how RAG task performance is affected when fewer gold documents are included in the context. (3) We show that saving retrieval time by decreasing approximate nearest neighbor (ANN) search accuracy in the retriever has only a minor effect on task performance. (4) We show that injecting noise into retrieval results in performance degradation. We find no setting that improves above the gold ceiling, contrary to a prior report \cite{Cuconasu_2024}. 

%% file: 2_rx-background.tex
\section{Background}
\label{sec:background}

A RAG pipeline is made up of two components: a retriever and a reader. The retriever component identifies relevant information from an exterior knowledge base which is included alongside a query in a prompt for the reader model \cite{yu2024evaluationretrievalaugmentedgenerationsurvey}. This process has been used as an effective alternative to expensive fine-tuning \cite{huang2023surveyhallucinationlargelanguage, meng2023locatingeditingfactualassociations} and is shown to reduce LLM hallucinations \cite{dale2022detectingmitigatinghallucinationsmachine}. 


\textbf{Retrieval models.} Dense vector embedding models have become the norm due to their improved performance above sparse retrievers that rely on metrics such as term frequency \cite{guoSemanticModelsFirststage2022}. These dense retrievers leverage nearest neighbor search algorithms to find document embeddings that are the closest to the query embedding. Of these dense models, most retrievers encode each document as a single vector \cite{zhaoDenseTextRetrieval2024}. However, multi-vector models that allow interactions between document terms and query terms such as ColBERT \cite{santhanamColBERTv2EffectiveEfficient2022} may generalize better to new datasets. In practical applications, most developers refer to text embedding leaderboards \cite{muennighoff2022mteb} or general information retrieval (IR) benchmarks such as BEIR \cite{thakur2021beir} to select a retriever.


\textbf{Approximate Nearest Neighbor (ANN) search.} 
Modern vector embeddings contain $\geq1024$ dimensions, resulting in severe search performance degradation (e.g., sifting through $\approx 170$GB of data for general knowledge corpora like Wikipedia) due to the curse of dimensionality. Consequently, RAG pipelines often employ approximate nearest neighbor search as a compromise, opting for faster search times at the expense of some accuracy \cite{Lewis_2020, Yoran_Wolfson_Ram_Berant_2024}. Despite this common practice, there is very little discussion in the literature regarding the optimal parameters for configuring ANN search, and the best way to balance the trade-off between accuracy and speed. Operating at a lower search accuracy could lead to massive improvements in search speed and memory footprint (for example, by eliminating the common re-ranking step~\cite[e.g.][]{tepper2023leanvec}). 



%% file: 3_rx-methods.tex
\section{Experiment setup}
\label{sec:experiments}

We conduct our experiments with two instruction-tuned LLMs: LLaMA (Llama-2-7b-chat) \cite{touvron2023llamaopenefficientfoundation} and Mistral (Mistral-7B-Instruct-v0.3) \cite{jiang2023mistral7b}. No further training or fine-tuning was performed. We avoided additional fine-tuning to ensure that our results are directly relevant to RAG pipelines currently being deployed across industry applications. Additional experiment details are in Appendix \ref{app:exp_setup}.

\textbf{Question answering (QA) and attributed QA.} For a query in a standard QA task, a RAG pipeline prompts an LLM to generate an answer based on information from a list of retrieved documents. In attributed QA, an LLM is also required to explicitly cite (e.g., by document ID) one or more of the documents used.  

\begin{figure}[ht]
    \centering
    \includegraphics[width=0.49\textwidth]{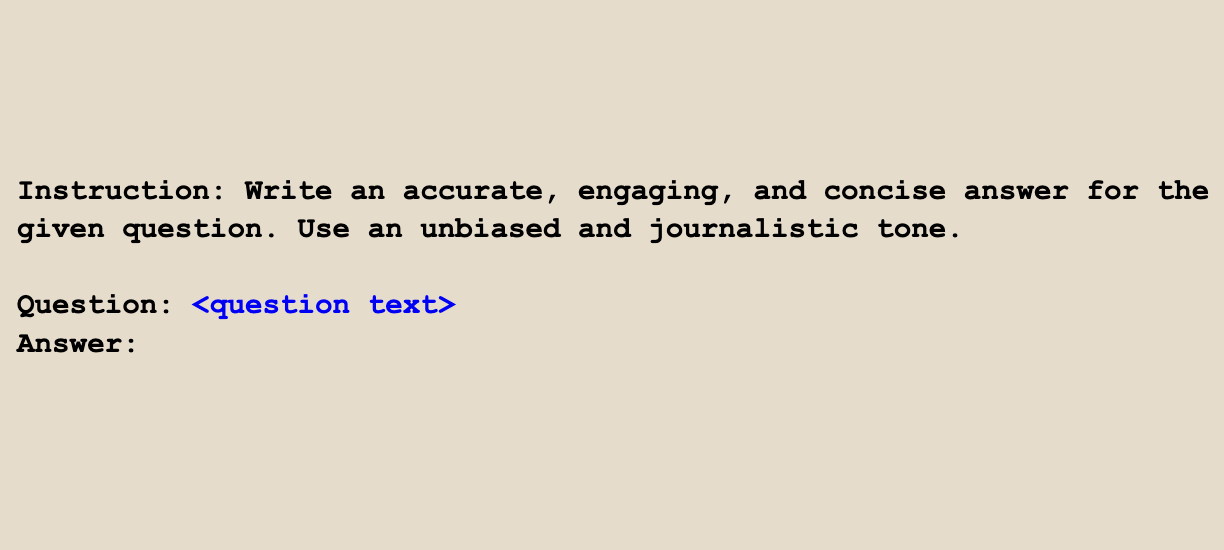}
    \includegraphics[width=0.49\textwidth]{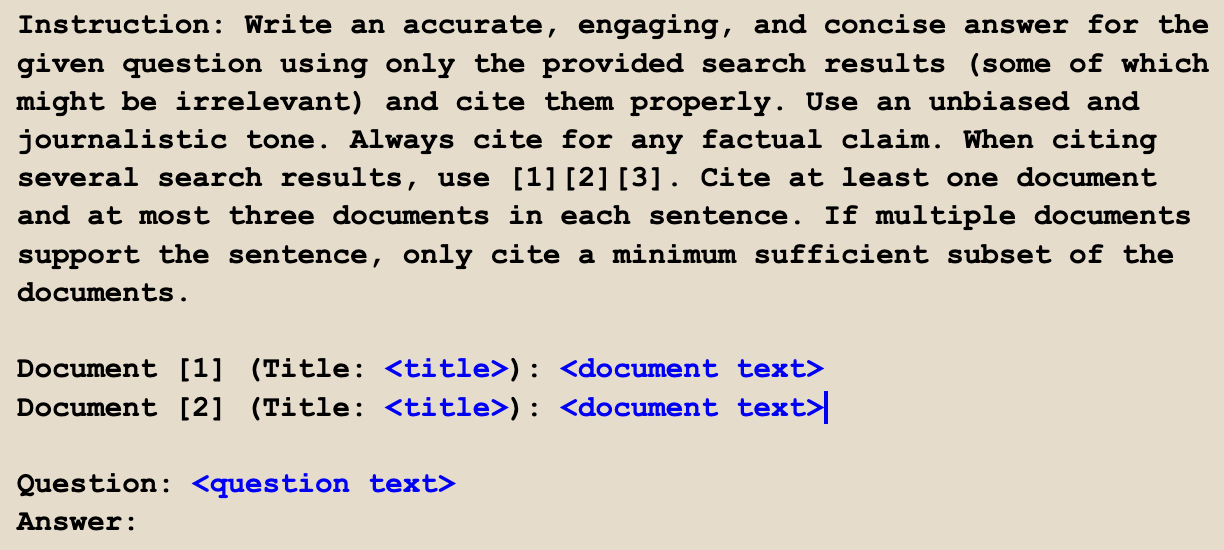}
    \caption{Example prompts for QA (left) and attributed QA (right, following \cite{Gao_Yen_Yu_Chen_2023}).}
    \label{fig:attributed_prompt}
\end{figure}

\textbf{Prompting.} Following previous work \cite{Gao_Yen_Yu_Chen_2023}, the models learn the desired format for attributing answers with citations via few-shot learning. We use 2-shot prompting for Mistral because of its longer context window and 1-shot prompting for LLaMA. We maintain the same prompt order for the experiments: system instruction, list of retrieved documents, then the query (see Figure \ref{fig:attributed_prompt}). When evaluating QA without attribution, 0-shots are given.

\subsection{Retrieval}

We chose to evaluate two high-performing, open-source dense retrieval models.
For single vector embeddings, we relied on \textbf{BGE-base} to embed documents (bge-base-en-v1.5 \cite{bge_embedding}, BEIR15 score of 0.533). We used the Intel SVS library\footnote{https://github.com/intel/ScalableVectorSearch} to search over these embeddings for efficient dense retrieval, exploiting its state-of-the-art graph-based ANN search performance \cite{aguerrebere_similarity_2023}. For multi-vector search, we used \textbf{ColBERTv2} \cite{DBLP:journals/corr/abs-2004-12832}, which leverages BERT embeddings to determine similarity between terms in documents and queries (BEIR15 score of 0.499).

\subsection{Datasets}
\label{sec:tasks}



\textbf{ASQA} is a long-form QA dataset for factoid questions designed to evaluate a model's performance on naturally-occurring ambiguous questions \cite{stelmakh-etal-2022-asqa}. It is made up of 948 queries and the ground truth documents are based on a 12/20/2018 Wikipedia dump with 21M passages.
We use the set of five gold documents provided by \cite{Gao_Yen_Yu_Chen_2023} which yields the best performance in their RAG pipeline. 

\textbf{QAMPARI} is an open-domain QA dataset in which the 1000 queries have several answers that can be found across multiple passages \cite{amouyal-etal-2023-qampari}. It is designed to be difficult for both retrieval and generation. As with ASQA, we use the five gold documents provided by \cite{Gao_Yen_Yu_Chen_2023} from the 2018 Wikipedia dump.

\textbf{Natural Questions (NQ)} is a dataset of 100k actual questions submitted to the Google search engine \cite{kwiatkowski-etal-2019-natural}. We follow \cite{Hsia_Shaikh_Wang_Neubig_2024} and use the KILT \cite{petroni2021kiltbenchmarkknowledgeintensive} version of the dataset, which consists of 2837 queries supported by 112M passages from a 2019 Wikipedia dump). It includes a short answer and at least one gold passage for each query. Though NQ has not traditionally corresponded to attributed QA, we adapt it to this task by simply prompting the language model to support statements with references to documents included in the context (see Figure \ref{fig:attributed_prompt}).




\subsection{Metrics} 
For retrieval, we report \textbf{recall@k}, which reflects the percentage of gold passages that have been retrieved in $k$ documents. We also refer to this as retriever recall or gold document recall. When using ANN, we also report \textbf{search recall@k}, that is the percentage of the $k$ exact nearest neighbors (according to the retriever similarity) that have been retrieved in the $k$ approximate nearest neighbors.

Correctness on the QA tasks is quantified by string \textbf{exact match recall (EM Rec.)}, or the percentage of short answers provided by the dataset which appear as exact substrings of the generated output. Note that for NQ, we report recall only over the top five gold answers following \cite{Gao_Yen_Yu_Chen_2023}.

To report citation quality, we use a process aligned with \cite{Rashkin_Nikolaev_Lamm_Aroyo_Collins_Das_Petrov_Tomar_Turc_Reitter_2023} that follows exactly the citation metrics found in the ALCE framework \cite{Gao_Yen_Yu_Chen_2023}: \textbf{citation recall} and \textbf{citation precision}. Citation recall is a measure of whether each generated statement includes citation(s) which entail it. Citation precision quantifies whether each individual citation is necessary to support a statement.

\textbf{Confidence intervals.} All metrics are computed for each query in the dataset, and averaged across all $n$ queries. To characterize the spread of the distribution, we compute 95\% confidence intervals (CIs) across queries using bootstrapping. That is, we resample $n$ queries with replacement from the true distribution, compute the mean, and repeat this process for 1000 bootstrap iterations. We then find the 2.5 and 97.5 percentiles for this distribution to yield the 95\% confidence intervals. Note that these bootstrapped CIs can be used to determine whether the difference between two distributions is statistically significant \cite{hesterbergBootstrap2011}.



%% file: 4_rx-results.tex
\section{Results}
\label{sec:results}

We first analyze how many retrieved documents should be included in the LLM context window to maximize correctness on the selected QA tasks. This is shown as a function of the number of retrieved nearest neighbors, $k$. Incorporating the retrieved documents narrows the performance disparity between the closed-book scenario ($k$=0) and the gold-document-only ceiling.
However, the performance of the evaluated retrieval systems still significantly lags behind the ideal. ColBERT usually outperforms BGE by a small margin.


\begin{figure}[h]
    \begin{center}
        \includegraphics[width=\textwidth]{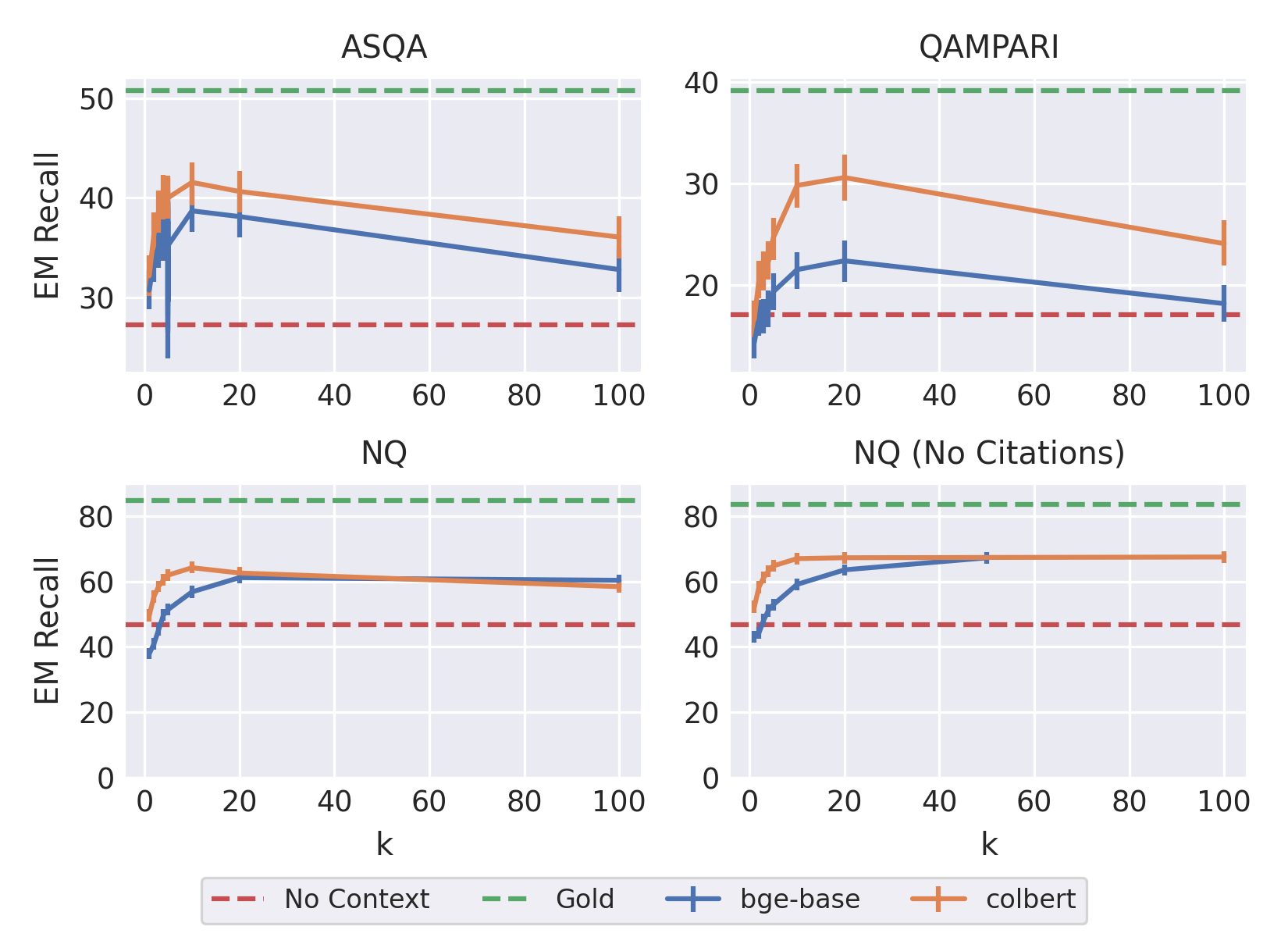}
    \end{center}
    \caption{Correctness achieved by Mistral with various numbers of documents retrieved with BGE-base and ColBERT. Optimal performance is observed with $k=10$ or $20$.}
    \label{fig:ndoc_mistral}
\end{figure}

\textbf{Correctness on QA begins to plateau around 5-10 documents.} We find that Mistral performs best for all three datasets with 10 or 20 documents (Figure \ref{fig:ndoc_mistral}). LLaMA performs best when $k=4$ or $5$ for ASQA and NQ, but $k=10$ for QAMPARI (Appendix Figure \ref{fig:ndoc_llama}). This difference between LLMs is likely due to LLaMA's shorter context window. We also find that adding the citation prompt to NQ results in almost no change to performance until $k > 10$. 
Tables \ref{tab:ndoc_asqa_mistral_bge} and \ref{tab:ndoc_asqa_mistral_colbert} show that citation recall generally peaks around the same point as QA correctness, while citation precision tends to peak at much lower $k$. Since citation precision measures how many of the cited documents are required for each statement, this suggests that showing the LLM more documents (i.e. at higher $k$) results in more extraneous, or unnecessary, citations. Citation measures for other datasets and models are in \ref{app:addl_ndoc}.

We further investigated where gold documents appear within the ranked list of retrieved documents. We found that gold documents typically ranked between 7-13th nearest neighbor (Appendix Table \ref{tab:median_gold_ranking}). Given these results, we conducted all subsequent analyses and experiments with 5-10 context documents, as these were generally good settings for QA performance even if some of the gold documents are missed. 



\begin{table}[h]
    \caption{Performance on ASQA with Mistral and various numbers of BGE-base retrieved documents, $k$, in the prompt. Optimal QA correctness is achieved at $k=10$, while it is $k=5$ for citation recall.}
    \centering
    \begin{tabular}{rrrrrrrr}
        \toprule
            \multicolumn{1}{c}{} & \multicolumn{1}{c}{\textbf{Ret.}} & \multicolumn{2}{c}{\textbf{EM Recall}} & \multicolumn{2}{c}{\textbf{Citation Recall}} & \multicolumn{2}{c}{\textbf{Citation Precision}} \\
            \cmidrule(r){2-2} 
            \cmidrule(r){3-4}
            \cmidrule(r){5-6}
            \cmidrule(r){7-8}
            k & \multicolumn{1}{c}{Rec@k} & \multicolumn{1}{c}{Mean} & \multicolumn{1}{c}{95\% CI} & \multicolumn{1}{c}{Mean} & \multicolumn{1}{c}{95\% CI} & \multicolumn{1}{c}{Mean} & \multicolumn{1}{c}{95\% CI} \\
            \midrule
            
           gold & 1 & 50.725 & \light{48.73 - 52.79} & 65.187 & \light{63.073 - 67.231} & 62.261 & \light{60.285 - 64.41} \\
            0 & 0 & 27.286 & \light{25.334 - 29.176} & - & - & - & - \\
            \midrule
            1 & 0.093 & 30.749 & \light{28.826 - 32.743} & 52.265 & \light{49.660 - 55.032} & 63.695 & \light{60.822 - 66.285} \\
            2 & 0.162 & 33.719 & \light{31.594 - 35.705} & 59.103 & \light{56.911 - 61.440} & \textbf{64.827} & \light{62.411 - 67.148} \\
            3 & 0.208 & 35.145 & \light{33.001 - 37.123} & 61.368 & \light{59.103 - 63.668} & 62.920 & \light{60.887 - 65.180} \\
            4 & 0.247 & 35.793 & \light{33.697 - 37.818} & 61.243 & \light{58.918 - 63.562} & 59.692 & \light{57.219 - 61.986} \\
            5 & 0.284 & 37.595 & \light{35.394 - 39.601} & \textbf{61.563} & \light{59.229 - 63.898} & 59.017 & \light{56.776 - 61.301} \\
            10 & 0.387 & \textbf{38.703} & \light{36.619 - 40.955} & 58.525 & \light{56.245 - 60.715} & 53.852 & \light{51.679 - 55.879} \\
            20 & 0.490 & 38.129 & \light{36.044 - 40.355} & 53.823 & \light{51.552 - 56.176} & 47.449 & \light{45.137 - 49.658} \\
            100 & 0.692 & 32.819 & \light{30.588 - 34.881} & 25.920 & \light{23.896 - 28.013} & 24.009 & \light{22.103 - 26.117} \\
           
            \bottomrule
    \end{tabular}
    \label{tab:ndoc_asqa_mistral_bge}
\end{table}

\begin{table}[h]
    \caption{Performance on ASQA with Mistral and various numbers of ColBERT retrieved documents, $k$, in the prompt. Optimal QA correctness is achieved at $k=10$, while it is $k=5$ for citation recall.}
    \centering
    \begin{tabular}{rrrrrrrr}
        \toprule
            \multicolumn{1}{c}{} & \multicolumn{1}{c}{\textbf{Ret.}} & \multicolumn{2}{c}{\textbf{EM Recall}} & \multicolumn{2}{c}{\textbf{Citation Recall}} & \multicolumn{2}{c}{\textbf{Citation Precision}} \\
            \cmidrule(r){2-2} 
            \cmidrule(r){3-4}
            \cmidrule(r){5-6}
            \cmidrule(r){7-8}
            k & \multicolumn{1}{c}{Rec@k} & \multicolumn{1}{c}{Mean} & \multicolumn{1}{c}{95\% CI} & \multicolumn{1}{c}{Mean} & \multicolumn{1}{c}{95\% CI} & \multicolumn{1}{c}{Mean} & \multicolumn{1}{c}{95\% CI} \\
            \midrule
            
            gold & 1 & 50.725 & \light{48.730 - 52.790} & 65.187 & \light{63.073 - 67.231} & 62.261 & \light{60.285 - 64.410} \\
            0 & 0 & 27.286 & \light{25.334 - 29.176} & - & - & - & - \\
            \midrule
            1 & 0.098 & 32.186 & \light{30.168 - 34.227} & 56.794 & \light{54.498 - 59.417} & \textbf{69.560} & \light{66.798 - 72.272} \\
            2 & 0.179 & 36.369 & \light{34.160 - 38.564} & 62.343 & \light{60.350 - 64.545} & 68.367 & \light{66.114 - 70.577} \\
            3 & 0.242 & 38.730 & \light{36.482 - 40.772} & 63.722 & \light{61.543 - 66.061} & 66.802 & \light{64.672 - 69.018} \\
            4 & 0.291 & 40.146 & \light{37.880 - 42.277} & 65.351 & \light{63.150 - 67.456} & 65.355 & \light{63.282 - 67.521} \\
            5 & 0.328 & 40.023 & \light{37.956 - 42.228} & \textbf{66.037} & \light{63.887 - 68.297} & 63.906 & \light{61.971 - 65.889} \\
            10 & 0.447 & \textbf{41.553} & \light{39.282 - 43.575} & 61.826 & \light{59.702 - 64.195} & 56.964 & \light{54.719 - 59.113} \\
            20 & 0.553 & 40.642 & \light{38.551 - 42.695} & 58.506 & \light{56.205 - 60.759} & 51.272 & \light{49.167 - 53.473} \\
            100 & 0.743 & 36.074 & \light{33.956 - 38.182} & 31.083 & \light{28.880 - 33.300} & 28.676 & \light{26.651 - 30.631} \\

            \bottomrule
    \end{tabular}
    \label{tab:ndoc_asqa_mistral_colbert}
\end{table}

\label{subsec:gold_analysis}
Based on the results above, we hypothesized that the ideal number of documents to include in a RAG pipeline is directly related to the number of gold documents that are retrieved within that $k$. This is relatively unexplored in the literature, as most have investigated how well LLMs can utilize the context and ignore non-gold documents \cite{Yoran_Wolfson_Ram_Berant_2024, Hsia_Shaikh_Wang_Neubig_2024, BehnamGhader_Miret_Reddy_2023, chenBenchmarkingLargeLanguage2024}. Because we observed similar trends across datasets, we dropped QAMPARI from the following results for simplicity. We re-analyze the results above for $k=10$ documents in the prompt, and simply bin the queries depending on the retriever recall (i.e. the percentage of retrieved gold documents).

\begin{figure}[ht!]
    \begin{center}
        \includegraphics[width=0.6\textwidth]{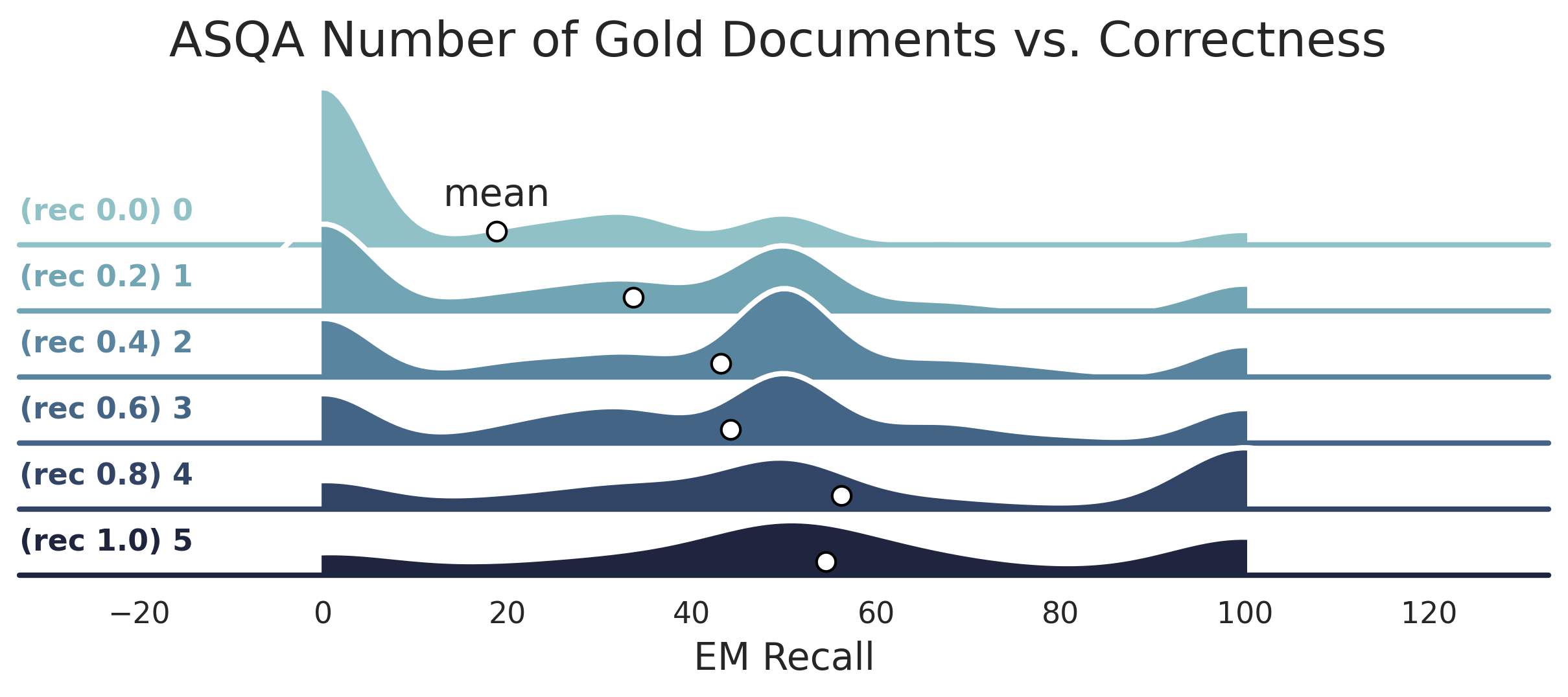}
    \end{center}
    \caption{The per-query relationship between the number of gold documents included in the prompt and the QA accuracy achieved with Mistral on ASQA. Including just one gold document significantly improves accuracy. There is a correlation between the number of gold documents and EM Recall.}
    \label{fig:recall_acc_mistral}
\end{figure}

\textbf{Including just one gold document highly increases correctness.} We observe a significant increase in the EM recall of queries with just one gold document in the prompt versus no gold documents. This is the case when either Mistral (Figure \ref{fig:recall_acc_mistral}) or LLaMA (Appendix Figure \ref{fig:recall_acc_llama}) is used as the reader module. We note that this trend was also observed in \cite{Hsia_Shaikh_Wang_Neubig_2024}.

\textbf{More gold documents correlates with higher correctness.} We find that increasing the number of gold documents in the prompt steadily increases QA correctness metrics. This is illustrated for Mistral in Figure \ref{fig:recall_acc_mistral} 
and LLaMA in Figure \ref{fig:recall_acc_llama}. 
We note that the difference in average correctness begins to plateau around a retrieval recall of 0.5. 
This supports the hypothesis that the ideal number of documents in the context window is directly related to the number of gold documents in that context window, in spite of the potential noise added by more non-gold documents. 



\subsection{Gold document recall and search accuracy regime}
\label{subsec:search_acc_regimes}
Next, we investigated how using approximate search affected RAG performance on the QA task. In particular, since the prior evidence suggests that gold documents are key to performance, we ran two sets of experiments to understand how both search recall and gold document recall affect QA performance.
First, we took the gold set and replaced some of these to reach a gold document recall target of 0.9, 0.7, or 0.5. For each query, we sampled a subset of the gold documents so that the average gold recall across all queries in a dataset reached the target, and populated the rest of the 10 documents with the nearest non-gold neighbors.
Second, we set the ANN search algorithm to achieve search recall targets of 0.95, 0.9, and 0.7 (details in Appendix ~\ref{app:tunning_ann}) and compared these to exact search (recall 1.0) for BGE-base. Figure \ref{fig:gold_search_recall} shows the results of these two experiments.

\textbf{Manipulating ANN search recall only results in minor decrements in QA performance.} We found that gold document recall (Fig. \ref{fig:gold_search_recall}, left) is a far bigger factor for QA performance than search recall (Fig. \ref{fig:gold_search_recall}, right). Setting the search recall@10 to 0.7 only results in a 2-3\% drop in gold document recall with respect to using exhaustive search (Table \ref{tab:search_regime_gold_recall}).
While our data is limited to a single dense retriever, it is the first experiment (to our knowledge) demonstrating that practitioners using current SOTA retrievers can take advantage of the speed and memory footprint benefits of ANN search with little to no adverse impact on RAG task performance.

\begin{figure}[h]
\centering
    \includegraphics{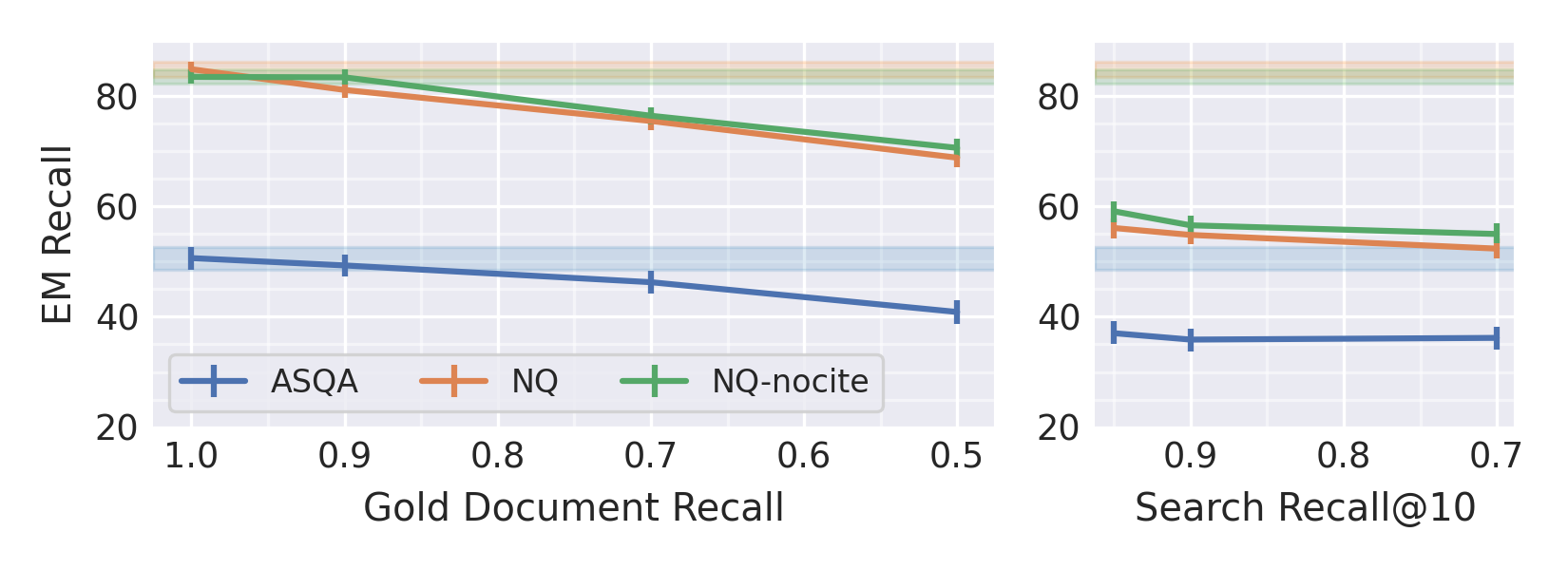}
    \caption{Gold document recall (left) has a greater impact on RAG QA performance compared to search recall (right). RAG pipeline uses Mistral and BGE-base. Shaded bar is ceiling performance using all gold documents per query. Error bars are 95\% bootstrap confidence intervals.}
    \label{fig:gold_search_recall_mistral}
\end{figure}

\begin{table}[h]
    \centering
    \caption{Gold document recall for the BGE-base retriever at different ANN search recall regimes. Reported as mean with 95\% confidence intervals (in grey).}
    \begin{tabular}{@{}cll@{}}
    \toprule
    \multirow{2}{*}{\textbf{ANN Search Recall@10}} & \multicolumn{2}{c}{\textbf{Doc. Recall@10}} \\ \cmidrule(l){2-3}
     & \multicolumn{1}{c}{ASQA} & \multicolumn{1}{c}{NQ} \\ \cmidrule(r){1-1} \cmidrule(l){2-2} \cmidrule(l){3-3}
    1.0 (exact) & 0.387 \hspace{2pt} \light{ 0.367 - 0.404} & 0.278 \hspace{2pt} \light{0.267 - 0.290} \\
    0.95 & 0.377 \hspace{2pt} \light{0.361 - 0.396} & 0.274 \hspace{2pt} \light{0.262 - 0.286} \\
    0.9 & 0.363 \hspace{2pt} \light{0.346 - 0.381} & 0.264 \hspace{2pt} \light{0.252 - 0.277} \\
    0.7 & 0.361 \hspace{2pt} \light{0.342 - 0.380} & 0.245 \hspace{2pt} \light{0.232 - 0.256} \\ 
    \bottomrule
    \end{tabular}
    \label{tab:search_regime_gold_recall}
\end{table}

\textbf{Citation metrics generally decrease as fewer supporting documents are available.} We observed that decreases in document recall and search recall lead to decreases in citation metrics (full results in Appendix \ref{app:addl_recall}). As with QA performance, decreases in document recall affect citation performance more than decreases in search recall (Table \ref{tab:citerec_varying_gold-search-recall}). However, this effect is less clear for the ASQA dataset, which is more likely to have multiple gold evidence documents that entail a single answer (\ref{app:addl_recall}).

\begin{table}[h]
    \centering
    \caption{Citation recall decreases as document recall and search recall decrease (NQ dataset, BGE-base retriever with Mistral reader). Values in parentheses are 95\% CIs.}
    \begin{tabular}{@{}cll@{}}
        \toprule
        & \multicolumn{2}{c}{\textbf{Citation Recall}}\\
        \midrule
        \multicolumn{1}{l}{\textbf{Doc. Recall@10}} & Mean & 95\% CI \\ \midrule
        1.0 & 75.041 & \light{73.772 - 76.397} \\
        0.9 & 72.084 & \light{70.765 - 73.409} \\
        0.7 & 66.669 & \light{65.258 - 68.148} \\
        0.5 & 60.827 & \light{59.428 - 62.292} \\
        \midrule
        \multicolumn{1}{l}{\textbf{Search Recall@10}} &  &  \\ \midrule
        0.95 & 55.340 & \light{53.872 - 56.814} \\
        0.9 & 52.443 & \light{50.835 - 53.972} \\
        0.7 & 49.856 & \light{48.405 - 51.340} \\ \bottomrule
    \end{tabular}
    \label{tab:citerec_varying_gold-search-recall}
\end{table}


\subsection{Injecting noisy documents of varying relevance}
Next, we explored whether the \textit{relevance} of the non-gold documents included in the context window affects the performance of the RAG pipeline on QA tasks. We define relevance as the similarity between the query and the retrieved document as defined by the corresponding retriever. A prior work \cite{Cuconasu_2024} made two claims about query-document similarity: (1) random non-gold documents increase QA performance above the gold-only ceiling; and (2) highly similar, non-gold documents are distracting and decrease QA performance.

To investigate claim 1, we added documents of varying similarity to either the gold set or the 5 most similar documents (nearest neighbor indices 0-4). First, we used BGE-base to retrieve all documents in the dataset for each ASQA query, assigning each neighbor a similarity score. We order the retrieved documents by this score and divide them into ten equal-sized bins. We define documents in the first bin 10\textsuperscript{th} percentile noise, the second bin 20\textsuperscript{th} percentile noise, etc. We randomly select 5 documents from each bin and append them to the prompt after either the gold or BGE-base retrieved documents. This setting follows the experiments in \cite{Cuconasu_2024}. Note that when injecting additional noise on top of the gold documents, the gold document recall (but not accuracy or F1) is still 1.0. 

\begin{figure}[h]
    \begin{center}
        \includegraphics[width=\textwidth]{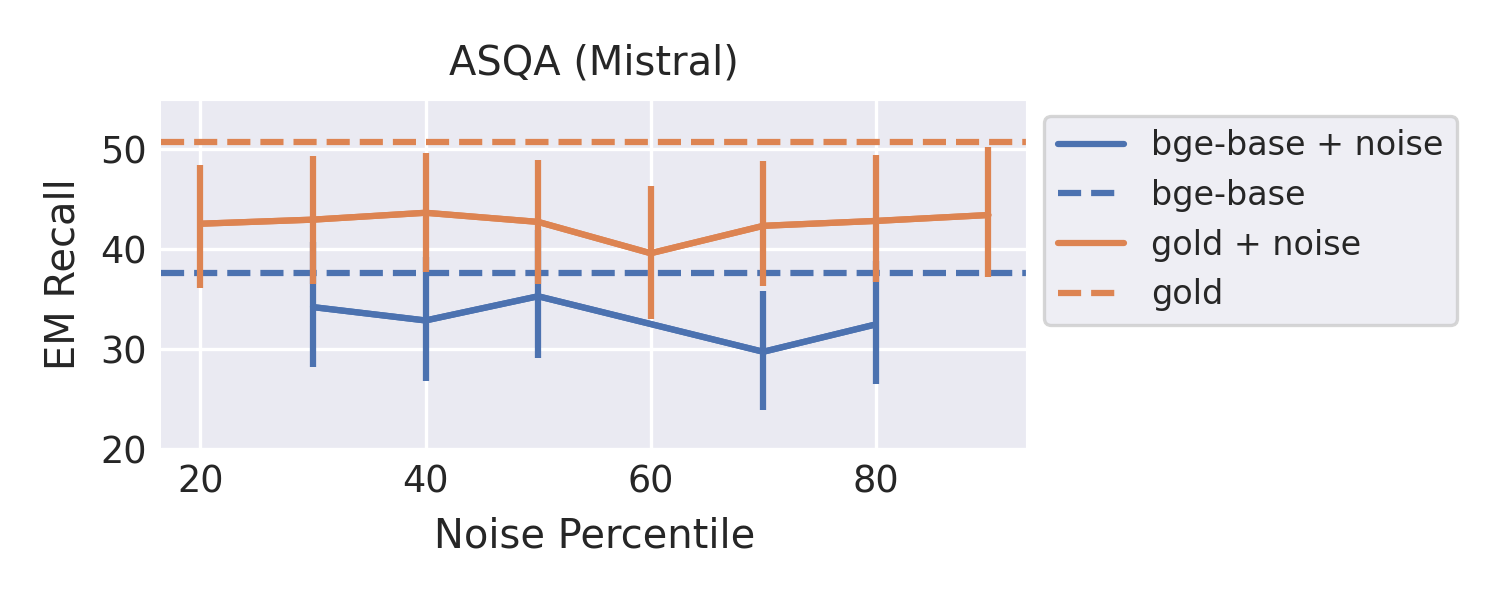}
    \end{center}
    \caption{ASQA Mistral performance after injecting noisy documents from various percentiles of similarity to the query. Adding noisy documents from all percentiles degrades QA correctness.}
    \label{fig:noise_percentile_asqa_mistral}
\end{figure}

\textbf{Our evidence does not clearly replicate claims in \cite{Cuconasu_2024} about injecting noise}. Contrary to claim 1, we find that adding noisy documents, regardless of their noise percentile, degrades correctness (Figure \ref{fig:noise_percentile_asqa_mistral}) and citation quality (\ref{app:addl_noise}) compared to the gold-only ceiling.

Figure \ref{fig:noise_percentile_asqa_mistral} also shows no consistent trend in performance changes with decreasingly similar documents. 
However, it is possible that claim 2 -- that very similar neighbors are more distracting than distantly similar neighbors -- might only be observed if we take the 1\textsuperscript{st} percentile of neighbors, as similarity is known to drop steeply with further neighbors (see Appendix Figure \ref{fig:all_neighbor_sim}). We therefore repeated a similar experiment with samples from the first 100 neighbors to test this claim.
We compare performance for Mistral on ASQA with 5 gold documents to performance when the $5^{\text{th}}-10^{\text{th}}$ or the $95^{\text{th}}-100^{\text{th}}$ nearest neighbors are added (Table \ref{tab:first100_asqa_mistral_gold}). Although QA performance still degrades, the effect is smaller---injecting more similar neighbors only drops performance by 1 point. Overall, injecting closer neighbors does not appear to be more detrimental than farther ones. Interestingly though, citation scores improve for farther neighbors. 
A similar pattern of QA performance was observed when using the same LLM as \cite{Cuconasu_2024} (Appendix \ref{app:addl_noise}).

These results are in line with \ref{subsec:search_acc_regimes}. Due to how ANN graph search is parameterized (\ref{app:tunning_ann}, lowering search recall adds \say{noisy} non-gold documents that are still similar to the query. Here and in \ref{subsec:search_acc_regimes}, we observed that injecting highly similar neighbors only mildly degrades downstream task performance.



\begin{table}[ht]
    \centering
    \caption{Mistral performance on ASQA when adding non-gold (\textit{noise}) documents based on their similarity ranking (between $5^{\text{th}}-100^{\text{th}}$ nearest neighbor).}
    \begin{tabular}{lllllll}
    \toprule
     \multirow{2}{*}{\textbf{Injected Noise}} & \multicolumn{2}{c}{\textbf{EM Recall}} & \multicolumn{2}{c}{\textbf{Citation Recall}} & \multicolumn{2}{c}{\textbf{Citation Precision}} \\ \cmidrule(r){2-3} \cmidrule(r){4-5} \cmidrule(r){6-7}
     & \multicolumn{1}{c}{Mean} & \multicolumn{1}{c}{95\% CI} & \multicolumn{1}{c}{Mean} & \multicolumn{1}{c}{95\% CI} & \multicolumn{1}{c}{Mean} & \multicolumn{1}{c}{95\% CI} \\  \cmidrule(r){1-1} \cmidrule(r){2-3} \cmidrule(r){4-5} \cmidrule(r){6-7}
    gold only & 50.73 & \light{48.73 - 52.79} & 65.19 & \light{63.07 - 67.23} & 62.26 & \light{60.29 - 64.41} \\
    gold + $5^{\text{th}}-10^{\text{th}}$ & 49.91 & \light{47.82 - 51.90} & 59.45 & \light{57.40 - 61.40} & 55.40 & \light{53.15 - 57.62} \\
    gold + $95^{\text{th}}-100^{\text{th}}$ & 49.24 & \light{47.05 - 51.46} & 58.69 & \light{56.37 - 61.11} & 56.14 & \light{53.96 - 58.23} \\ \bottomrule
    \end{tabular}
    \label{tab:first100_asqa_mistral_gold}
\end{table}

%% file: 5_rx-conclusion.tex
\section{Conclusion}
\label{sec:conclusion}
Overall, our experiments suggest that models that can retrieve a higher number of gold documents will maximize QA performance. We also observe that leveraging ANN search to retrieve documents with a lower recall results in only slight QA performance degradation, which correlates with the very minor changes to gold document recall. Thus, operating at a lower search recall regime is a viable option in practice to potentially increase speed and memory efficiency. We also find that, contrary to a prior study \cite{Cuconasu_2024}, injecting noisy documents alongside gold or retrieved documents degrades correctness compared to the gold ceiling. We also find that it has an inconsistent effect on citation metrics. This suggests that the impact of document noise on RAG performance requires further study. 

Future work should test the generality of these findings in other settings. Understanding how approximate vs. exact search affects multi-vector retrievers such as \cite{santhanamColBERTv2EffectiveEfficient2022} would be interesting, especially given their generally good performance. 
Additionally, we only evaluated systems where the retriever and reader are trained separately. RAG systems trained end-to-end (e.g. Fusion-in-Decoder (FiD) \cite{izacardLeveragingPassageRetrieval2021} model), may rely less on gold documents such that retrieval metrics are not useful relevance markers \cite{Salemi_Zamani_2024}.

%% file: 6_rx-appendix.tex
\section{Appendix}

\subsection{Experiment setup}
\label{app:exp_setup}

We conduct the retrieval portion of the experiments on a 2-socket 2nd generation Intel{\textregistered} Xeon{\textregistered} 8280L @2.70GHz CPUs with 28 cores (2x hyperthreading enabled) and 384GB DDR4 memory (@2933MT/s) per socket, running Ubuntu 22.04. Retrieval results were saved to files and were inserted into the prompt (Figure \ref{fig:attributed_prompt}) for the LLM during the reader portion of the experiments. 

We ran LLM inference on NVIDIA GPUs of varying models (NVIDIA Titan Xp or X Pascal series, or NVIDIA A40). The CPU hosts for these GPU nodes were either Intel{\textregistered} E5-2699 v4 or Intel{\textregistered} Xeon{\textregistered} 8280 or 8280L.
Run time for generating answers for all queries in ASQA and QAMPARI was approximately 30 minutes for Mistral and 20 minutes for LLaMA. Because there are more queries in NQ, run time was approximately 1.5 hours for Mistral and 1 hour for LLaMA.

A temperature of 1 and top p of 0.95 were used for generation with both LLMs.

\subsubsection{Tuning ANN search}
\label{app:tunning_ann}
Approximate nearest neighbor (ANN) techniques that utilize graphs are notable for their exceptional search accuracy and speed, particularly with data of high dimensionality \cite{li_approximate_2020,malkov_efficient_2020,aguerrebere_similarity_2023}. We leverage the Intel SVS library's graph-based search capabilities to our advantage. These graph-based approaches employ proximity graphs, in which the nodes correspond to data vectors. A connection is established between two nodes if they meet a specific property or neighborhood criterion, leveraging the natural structure found within the data.

The search process begins at a predetermined starting node and progresses through the graph, moving from one node to the next, each step bringing the search closer to the nearest neighbor by following a best-first search strategy. To prevent becoming trapped in a local minimum and to enable the discovery of multiple nearest neighbors, backtracking is employed \cite{malkov_efficient_2020,subramanya_diskann_2019}. Increasing the extent of backtracking means that a larger section of the graph is examined, which enhances the precision of the search but also results in a longer and therefore slower process. By adjusting the setting that determines the level of backtracking, we can fine-tune the balance between search accuracy, reflected in the quality of the nearest neighbors found, and the number of queries that can be handled per second. The Intel SVS library uses the \texttt{search\_window\_size} parameter to set the search accuracy vs. speed trade-off. By changing the \texttt{search\_window\_size} we set the retrieval module to operate at different search recall regimes.

\subsection{Nearest neighbor similarity}
\begin{figure}[ht]
    \centering
    \subfloat[Neighbor index (x-axis) shown linearly.]{\includegraphics[width=0.5\textwidth]{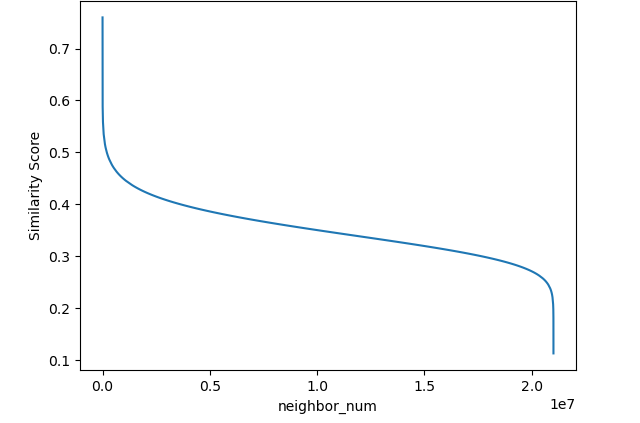}}
    \subfloat[Neighbor index (x-axis) shown on a log-scale.]{\includegraphics[width=0.5\textwidth]{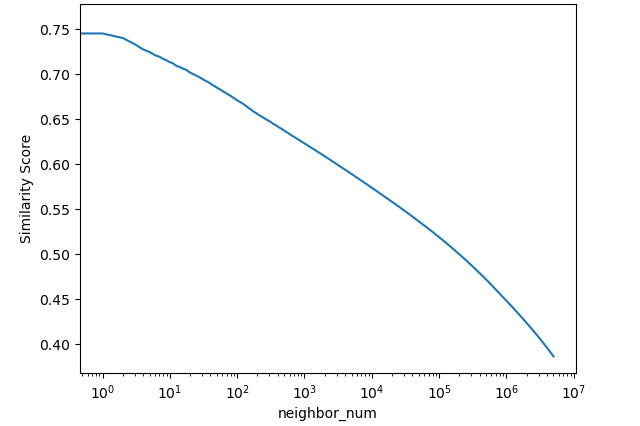}}
    \caption{Average similarity of nearest neighbors for the ALCE dataset using BGE-base as a retriever.}
    \label{fig:all_neighbor_sim}
\end{figure}

\subsection{Gold documents as nearest neighbors}
\begin{figure}[ht]
    \begin{center}
        \includegraphics[width=\textwidth]{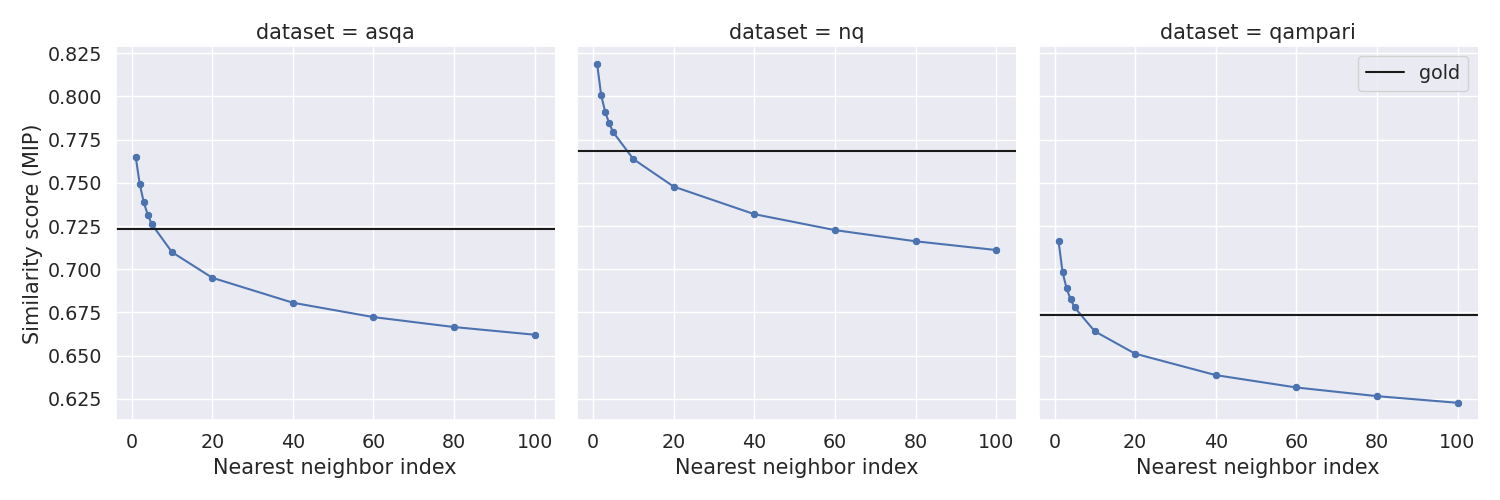}
    \end{center}
    \caption{The similarity score (maximum inner product) of BGE-base neighbors, averaged across queries within the dataset. The solid black line is the mean similarity score for gold documents (grand mean first within query then across queries).}
    \label{fig:avg_sim_scores}
\end{figure}

Table \ref{tab:median_gold_ranking} shows how gold documents rank in the nearest neighbors. The 25th, 50th, and 75th percentiles are provided. Figure \ref{fig:avg_sim_scores} shows how the average similarity score of gold documents, compared to the average similarity of different neighbor rankings. 

\begin{table}[ht]
    \centering
    \caption{Nearest neighbor ranking of gold documents for each retriever and dataset. Since these distributions are skewed across queries, we report quartiles (1st, 2nd / median, and 3rd). The median neighbor ranking of gold documents is between 7 and 13.}
    \begin{tabular}{@{}lllll@{}}
    \toprule
     &  & Q1 (25) & Q2 (50) & Q3 (75) \\ \midrule
    \multirow{3}{*}{ASQA} & BGE-base & 3.0 & 8.0 & 25.0 \\
     & BGE-large & 3.0 & 8.0 & 25.0 \\
     & ColBERTv2 & 3.0 & 7.0 & 21.0 \\ \midrule
    \multirow{3}{*}{NQ} & BGE-base & 3.0 & 11.0 & 32.0 \\
     & BGE-large & 3.0 & 10.5 & 33.0 \\
     & ColBERTv2 & 2.0 & 7.0 & 24.0 \\ \midrule
    \multirow{3}{*}{QAMPARI} & BGE-base & 4.0 & 13.0 & 37.0 \\
     & BGE-large & 4.0 & 13.0 & 41.0 \\
     & ColBERTv2 & 4.0 & 11.0 & 31.0 \\ \bottomrule
    \end{tabular}
    \label{tab:median_gold_ranking}
\end{table}

\pagebreak

\subsection{Additional varied number of neighbors results}
\label{app:addl_ndoc}

\begin{figure}[ht]
    \begin{center}
        \includegraphics[width=\textwidth]{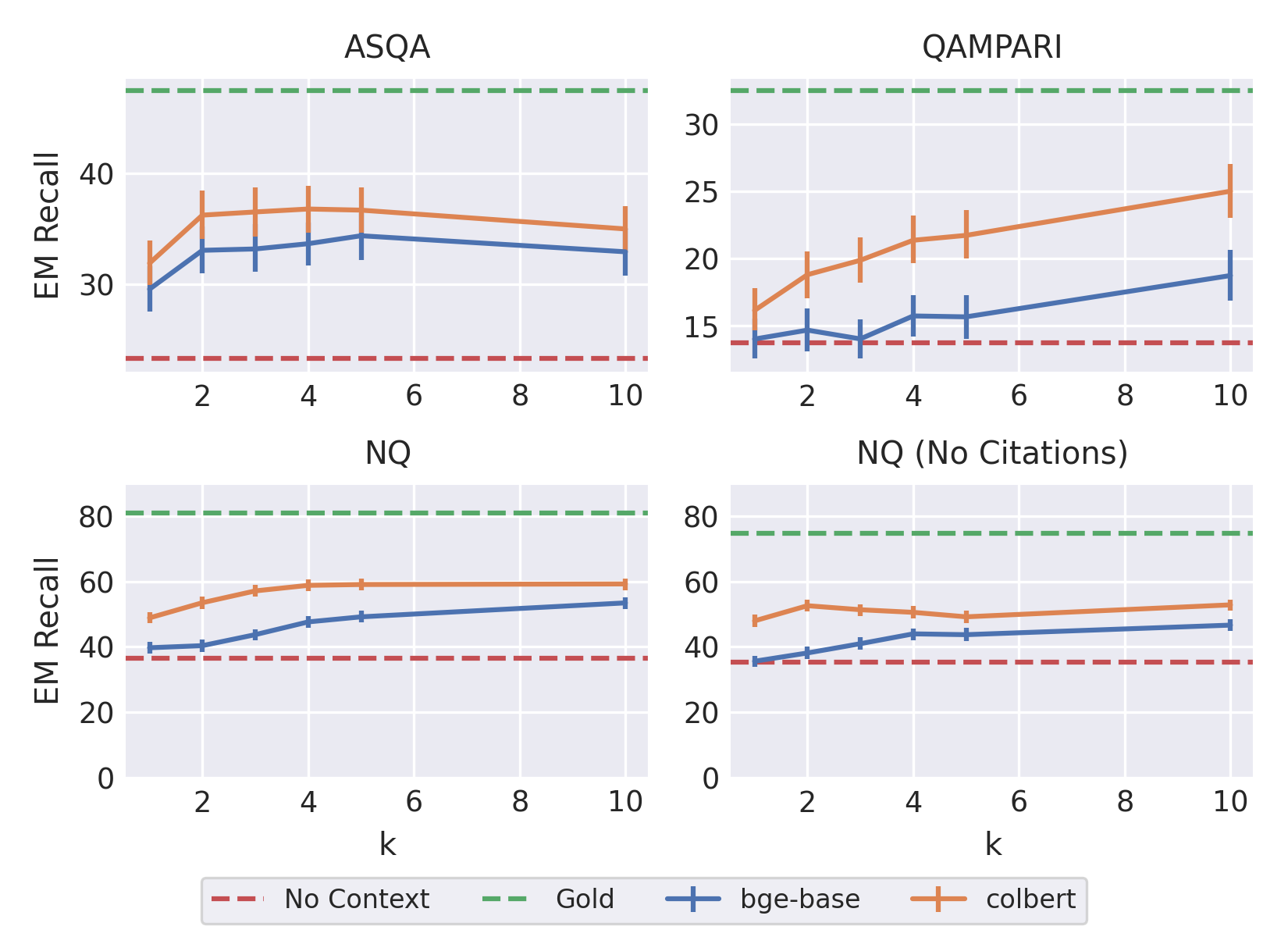}
        
    \end{center}
    \caption{Correctness achieved by prompting LLaMA with various numbers of documents retrieved with BGE-base and ColBERT, \textit{k}, included in the prompts. Optimal performance is observed with $k=4$ or $5$ for ASQA and NQ, while optimal performance for QAMPARI is achieved with $k=10$.}
    \label{fig:ndoc_llama}
\end{figure}

Correctness results for all three datasets with LLaMA are in Figure \ref{fig:ndoc_llama}. We present detailed results for including varied number of retrieved documents, $k$, for ASQA with LLaMA in Tables \ref{tab:ndoc_asqa_llama_bge} and \ref{tab:ndoc_asqa_llama_colbert}. Detailed results on NQ are shown in Tables \ref{tab:ndoc_nq_mistral_bge}, \ref{tab:ndoc_nq_mistral_colbert}, \ref{tab:ndoc_asqa_llama_bge} and \ref{tab:ndoc_nq_llama_colbert}. Detailed results on QAMPARI are included in Tables \ref{tab:ndoc_qampari_mistral_bge}, \ref{tab:ndoc_qampari_mistral_colbert}, \ref{tab:ndoc_qampari_llama_bge} and \ref{tab:ndoc_qampari_llama_colbert}.


\begin{table}[ht]
    \centering
    \caption{Correctness and citation quality on ASQA achieved with LLaMA with various numbers of BGE-base retrieved documents, $k$, included in the prompt.}
    \begin{tabular}{rrrrrrrr}
        \toprule
             \multicolumn{1}{c}{} & \multicolumn{1}{c}{\textbf{Ret.}} & \multicolumn{2}{c}{\textbf{EM Recall}} & \multicolumn{2}{c}{\textbf{Citation Recall}} & \multicolumn{2}{c}{\textbf{Citation Precision}} \\
            \cmidrule(r){2-2} 
            \cmidrule(r){3-4}
            \cmidrule(r){5-6}
            \cmidrule(r){7-8}
            k & \multicolumn{1}{c}{Rec@k} & \multicolumn{1}{c}{Mean} & \multicolumn{1}{c}{95\% CI} & \multicolumn{1}{c}{Mean} & \multicolumn{1}{c}{95\% CI} & \multicolumn{1}{c}{Mean} & \multicolumn{1}{c}{95\% CI} \\
            \midrule

            \multicolumn{1}{l}{gold} & 1 & 47.466 & \light{45.304 - 49.426} & 46.326 & \light{44.123 - 48.524} & 46.294 & \light{44.082 - 48.686} \\
            0 & 0 & 23.327 & \light{21.355 - 25.252} & \multicolumn{1}{r}{-} & - & \multicolumn{1}{r}{-} & - \\
            1 & 0.093 & 29.587 & \light{27.553 - 31.639} & 28.025 & \light{25.854 - 30.463} & 35.261 & \light{32.479 - 38.025} \\
            \midrule
            2 & 0.162 & 33.09 & \light{31.052 - 35.208} & 47.128 & \light{44.734 - 49.398} & 53.155 & \light{50.567 - 55.656} \\
            3 & 0.208 & 33.212 & \light{31.144 - 35.472} & 50.786 & \light{48.408 - 53.057} & 52.383 & \light{49.939 - 54.661} \\
            4 & 0.247 & 33.686 & \light{31.706 - 35.607} & 46.221 & \light{43.969 - 48.433} & 47.776 & \light{45.412 - 50.114} \\
            5 & 0.284 & 34.402 & \light{32.205 - 36.468} & 42.227 & \light{40.019 - 44.503} & 41.97 & \light{39.805 - 44.187} \\
            10 & 0.387 & 32.956 & \light{30.789 - 35.019} & 34.6 & \light{32.18 - 37.153} & 30.585 & \light{28.381 - 32.625} \\
           
            \bottomrule
    \end{tabular}
    \label{tab:ndoc_asqa_llama_bge}
\end{table}

\begin{table}[ht]
    \centering
    \caption{Correctness and citation quality on ASQA achieved with LLaMA with various numbers of ColBERT retrieved documents, $k$, included in the prompt.}
    \begin{tabular}{rrrrrrrr}
        \toprule
             \multicolumn{1}{c}{} & \multicolumn{1}{c}{\textbf{Ret.}} & \multicolumn{2}{c}{\textbf{EM Recall}} & \multicolumn{2}{c}{\textbf{Citation Recall}} & \multicolumn{2}{c}{\textbf{Citation Precision}} \\
            \cmidrule(r){2-2} 
            \cmidrule(r){3-4}
            \cmidrule(r){5-6}
            \cmidrule(r){7-8}
            k & \multicolumn{1}{c}{Rec@k} & \multicolumn{1}{c}{Mean} & \multicolumn{1}{c}{95\% CI} & \multicolumn{1}{c}{Mean} & \multicolumn{1}{c}{95\% CI} & \multicolumn{1}{c}{Mean} & \multicolumn{1}{c}{95\% CI} \\
            \midrule

            gold & 1 & 47.466 & \light{45.304 - 49.426} & 46.326 & \light{44.123 - 48.524} & 46.294 & \light{44.082 - 48.686} \\
            0 & 0 & 23.327 & \light{21.355 - 25.252} & - & - & - & - \\
            \midrule
            1 & 0.098 & 31.928 & \light{29.979 - 33.962} & 34.944 & \light{32.569 - 37.116} & 43.637 & \light{41.007 - 46.503} \\
            2 & 0.179 & 36.262 & \light{34.089 - 38.509} & 52.282 & \light{49.951 - 54.57} & 58.651 & \light{56.199 - 60.848} \\
            3 & 0.242 & 36.548 & \light{34.33 - 38.787} & 54.545 & \light{52.246 - 56.676} & 57.067 & \light{54.78 - 59.45} \\
            4 & 0.291 & 36.813 & \light{34.688 - 38.9} & 50.63 & \light{48.394 - 52.85} & 51.505 & \light{49.236 - 53.808} \\
            5 & 0.328 & 36.712 & \light{34.668 - 38.79} & 46.293 & \light{43.897 - 48.806} & 44.565 & \light{42.292 - 46.945} \\
            10 & 0.447 & 35.016 & \light{32.844 - 37.057} & 37.334 & \light{34.937 - 39.771} & 32.033 & \light{29.975 - 34.319} \\

            \bottomrule
    \end{tabular}
    \label{tab:ndoc_asqa_llama_colbert}
\end{table}

\begin{table}[ht]
    \centering
    \caption{Correctness and citation quality on NQ achieved with Mistral with various numbers of BGE-base retrieved documents, $k$, included in the prompt.}
    \begin{tabular}{rrrrrrrr}
        \toprule
             \multicolumn{1}{c}{} & \multicolumn{1}{c}{\textbf{Ret.}} & \multicolumn{2}{c}{\textbf{EM Recall}} & \multicolumn{2}{c}{\textbf{Citation Recall}} & \multicolumn{2}{c}{\textbf{Citation Precision}} \\
            \cmidrule(r){2-2} 
            \cmidrule(r){3-4}
            \cmidrule(r){5-6}
            \cmidrule(r){7-8}
            k & \multicolumn{1}{c}{Rec@k} & \multicolumn{1}{c}{Mean} & \multicolumn{1}{c}{95\% CI} & \multicolumn{1}{c}{Mean} & \multicolumn{1}{c}{95\% CI} & \multicolumn{1}{c}{Mean} & \multicolumn{1}{c}{95\% CI} \\
            \midrule

          gold & 1 & 84.646 & \light{83.222 - 86.042} & 74.66 & \light{73.349 - 76.006} & 64.407 & \light{63.179 - 65.615} \\
        0 &  & 46.696 & \light{44.836 - 48.467} & - & - & - & - \\
        \midrule
        1 & 0.072 & 37.871 & \light{36.164 - 39.726} & 31.101 & \light{29.567 - 32.776} & 36.4 & \light{34.695 - 38.181} \\
        2 & 0.117 & 40.925 & \light{39.02 - 42.792} & 43.51 & \light{41.923 - 45.18} & 39.554 & \light{38.067 - 41.129} \\
        3 & 0.152 & 45.301 & \light{43.497 - 47.127} & 47.423 & \light{45.796 - 48.944} & 40.574 & \light{39.101 - 42.005} \\
        4 & 0.181 & 49.735 & \light{47.902 - 51.639} & 49.943 & \light{48.38 - 51.534} & 43.404 & \light{42.007 - 44.826} \\
        5 & 0.205 & 51.421 & \light{49.559 - 53.226} & 51.062 & \light{49.51 - 52.51} & 43.69 & \light{42.326 - 45.13} \\
        10 & 0.279 & 56.778 & \light{54.916 - 58.654} & 55.869 & \light{54.395 - 57.241} & 46.267 & \light{45.014 - 47.542} \\
        20 & 0.355 & 61.145 & \light{59.391 - 62.883} & 57.173 & \light{55.754 - 58.595} & 46.497 & \light{45.282 - 47.711} \\
        100 & 0.53 & 60.333 & \light{58.548 - 62.143} & 49.816 & \light{48.332 - 51.397} & 36.671 & \light{35.502 - 37.921} \\

            \bottomrule
    \end{tabular}
    \label{tab:ndoc_nq_mistral_bge}
\end{table}

\begin{table}[ht]
    \centering
    \caption{Correctness and citation quality on NQ achieved with Mistral with various numbers of ColBERT retrieved documents, $k$, included in the prompt.}
    \begin{tabular}{rrrrrrrr}
        \toprule
             \multicolumn{1}{c}{} & \multicolumn{1}{c}{\textbf{Ret.}} & \multicolumn{2}{c}{\textbf{EM Recall}} & \multicolumn{2}{c}{\textbf{Citation Recall}} & \multicolumn{2}{c}{\textbf{Citation Precision}} \\
            \cmidrule(r){2-2} 
            \cmidrule(r){3-4}
            \cmidrule(r){5-6}
            \cmidrule(r){7-8}
            k & \multicolumn{1}{c}{Rec@k} & \multicolumn{1}{c}{Mean} & \multicolumn{1}{c}{95\% CI} & \multicolumn{1}{c}{Mean} & \multicolumn{1}{c}{95\% CI} & \multicolumn{1}{c}{Mean} & \multicolumn{1}{c}{95\% CI} \\
            \midrule

        gold & 1 & 84.646 & \light{83.222 - 86.042} & 74.66 & \light{73.349 - 76.006} & 64.407 & \light{63.179 - 65.615} \\
        0 & \multicolumn{1}{l}{} & 46.696 & \light{44.836 - 48.467} & - & - & - & - \\
        \midrule
        1 & 0.122 & 49.567 & \light{47.797 - 51.463} & 67.754 & \light{66.276 - 69.245} & 77.295 & \light{75.828 - 78.863} \\
        2 & 0.179 & 55.398 & \light{53.542 - 57.244} & 72.288 & \light{70.925 - 73.617} & 68.703 & \light{67.417 - 69.971} \\
        3 & 0.214 & 58.486 & \light{56.714 - 60.204} & 71.02 & \light{69.784 - 72.254} & 63.799 & \light{62.58 - 65.048} \\
        4 & 0.237 & 60.41 & \light{58.688 - 62.178} & 69.883 & \light{68.64 - 71.088} & 60.987 & \light{59.797 - 62.236} \\
        5 & 0.254 & 61.916 & \light{60.134 - 63.66} & 68.635 & \light{67.291 - 69.896} & 59.862 & \light{58.683 - 61.101} \\
        10 & 0.321 & 64.171 & \light{62.424 - 66.056} & 67.233 & \light{65.901 - 68.508} & 55.348 & \light{54.1 - 56.477} \\
        20 & 0.381 & 62.558 & \light{60.874 - 64.434} & 64.701 & \light{63.314 - 66.002} & 50.239 & \light{49.051 - 51.368} \\
        100 & 0.506 & 58.384 & \light{56.502 - 60.204} & 50.21 & \light{48.575 - 51.79} & 35.233 & \light{34.014 - 36.53} \\

            \bottomrule
    \end{tabular}
    \label{tab:ndoc_nq_mistral_colbert}
\end{table}

\begin{table}[ht]
    \centering
    \caption{Correctness and citation quality on NQ achieved with LLaMA with various numbers of BGE-base retrieved documents, $k$, included in the prompt.}
    \begin{tabular}{rrrrrrrr}
        \toprule
             \multicolumn{1}{c}{} & \multicolumn{1}{c}{\textbf{Ret.}} & \multicolumn{2}{c}{\textbf{EM Recall}} & \multicolumn{2}{c}{\textbf{Citation Recall}} & \multicolumn{2}{c}{\textbf{Citation Precision}} \\
            \cmidrule(r){2-2} 
            \cmidrule(r){3-4}
            \cmidrule(r){5-6}
            \cmidrule(r){7-8}
            k & \multicolumn{1}{c}{Rec@k} & \multicolumn{1}{c}{Mean} & \multicolumn{1}{c}{95\% CI} & \multicolumn{1}{c}{Mean} & \multicolumn{1}{c}{95\% CI} & \multicolumn{1}{c}{Mean} & \multicolumn{1}{c}{95\% CI} \\
            \midrule

            gold & 1 & 80.932 & \light{79.52 - 82.341} & 54.725 & \light{53.288 - 56.182} & 55.763 & \light{54.364 - 57.225} \\
            0 & 0 & 36.389 & \light{34.544 - 38.174} & - & - & - & - \\
            \midrule
            1 & 0.072 & 39.715 & \light{37.821 - 41.559} & 23.417 & \light{22.007 - 24.828} & 27.726 & \light{26.215 - 29.241} \\
            2 & 0.117 & 40.357 & \light{38.421 - 42.192} & 33.892 & \light{32.343 - 35.487} & 35.623 & \light{34.07 - 37.219} \\
            3 & 0.152 & 43.718 & \light{41.91 - 45.436} & 36.163 & \light{34.775 - 37.658} & 35.104 & \light{33.599 - 36.504} \\
            4 & 0.181 & 47.611 & \light{45.858 - 49.384} & 35.693 & \light{34.254 - 37.059} & 34.527 & \light{33.198 - 35.826} \\
            5 & 0.205 & 49.159 & \light{47.374 - 50.969} & 34.839 & \light{33.47 - 36.317} & 32.967 & \light{31.56 - 34.304} \\
            10 & 0.279 & 53.424 & \light{51.638 - 55.164} & 28.622 & \light{27.287 - 29.967} & 25.923 & \light{24.689 - 27.058} \\
            20 & 0.381 & 62.558 & \light{60.874 - 64.434} & 64.701 & \light{63.314 - 66.002} & 50.239 & \light{49.051 - 51.368} \\
            100 & 0.506 & 58.384 & \light{56.502 - 60.204} & 50.21 & \light{48.575 - 51.79} & 35.233 & \light{34.014 - 36.53} \\

            \bottomrule
    \end{tabular}
    \label{tab:ndoc_nq_llama_bge}
\end{table}

\begin{table}[ht]
    \centering
    \caption{Correctness and citation quality on NQ achieved with LLaMA with various numbers of ColBERT retrieved documents, $k$, included in the prompt.}
    \begin{tabular}{rrrrrrrr}
        \toprule
             \multicolumn{1}{c}{} & \multicolumn{1}{c}{\textbf{Ret.}} & \multicolumn{2}{c}{\textbf{EM Recall}} & \multicolumn{2}{c}{\textbf{Citation Recall}} & \multicolumn{2}{c}{\textbf{Citation Precision}} \\
            \cmidrule(r){2-2} 
            \cmidrule(r){3-4}
            \cmidrule(r){5-6}
            \cmidrule(r){7-8}
            k & \multicolumn{1}{c}{Rec@k} & \multicolumn{1}{c}{Mean} & \multicolumn{1}{c}{95\% CI} & \multicolumn{1}{c}{Mean} & \multicolumn{1}{c}{95\% CI} & \multicolumn{1}{c}{Mean} & \multicolumn{1}{c}{95\% CI} \\
            \midrule

            gold & 1 & 80.932 & \light{79.52 - 82.341} & 54.725 & \light{53.288 - 56.182} & 55.763 & \light{54.364 - 57.225} \\
            0 & 0 & 36.389 & \light{34.544 - 38.174} & - & - & - & - \\
            1 & 0.122 & 48.847 & \light{47.126 - 50.617} & 54.977 & \light{53.446 - 56.483} & 67.286 & \light{65.586 - 68.933} \\
            2 & 0.179 & 53.481 & \light{51.638 - 55.306} & 57.519 & \light{56.141 - 58.831} & 65.442 & \light{64.0 - 66.948} \\
            3 & 0.214 & 57.088 & \light{55.27 - 58.972} & 56.378 & \light{54.94 - 57.854} & 59.892 & \light{58.342 - 61.326} \\
            4 & 0.237 & 58.786 & \light{57.031 - 60.592} & 48.572 & \light{47.112 - 50.152} & 51.234 & \light{49.727 - 52.642} \\
            5 & 0.254 & 59.04 & \light{57.208 - 60.804} & 44.974 & \light{43.704 - 46.308} & 46.73 & \light{45.352 - 48.104} \\
            10 & 0.321 & 59.212 & \light{57.349 - 60.945} & 20.997 & \light{19.857 - 22.213} & 21.104 & \light{19.948 - 22.278} \\
            20 & 0.381 & 62.558 & \light{60.874 - 64.434} & 64.701 & \light{63.314 - 66.002} & 50.239 & \light{49.051 - 51.368} \\
            100 & 0.506 & 58.384 & \light{56.502 - 60.204} & 50.21 & \light{48.575 - 51.79} & 35.233 & \light{34.014 - 36.53} \\

            \bottomrule
    \end{tabular}
    \label{tab:ndoc_nq_llama_colbert}
\end{table}

\begin{table}[ht]
    \centering
    \caption{Correctness and citation quality on QAMPARI achieved with Mistral with various numbers of BGE-base retrieved documents, $k$, included in the prompt.}
    \begin{tabular}{rrrrrrrr}
        \toprule
             \multicolumn{1}{c}{} & \multicolumn{1}{c}{\textbf{Ret.}} & \multicolumn{2}{c}{\textbf{EM Recall}} & \multicolumn{2}{c}{\textbf{Citation Recall}} & \multicolumn{2}{c}{\textbf{Citation Precision}} \\
            \cmidrule(r){2-2} 
            \cmidrule(r){3-4}
            \cmidrule(r){5-6}
            \cmidrule(r){7-8}
            k & \multicolumn{1}{c}{Rec@k} & \multicolumn{1}{c}{Mean} & \multicolumn{1}{c}{95\% CI} & \multicolumn{1}{c}{Mean} & \multicolumn{1}{c}{95\% CI} & \multicolumn{1}{c}{Mean} & \multicolumn{1}{c}{95\% CI} \\
            \midrule

            gold & 1 & 39.152 & \light{36.879 - 41.44} & 23.239 & \light{20.827 - 25.708} & 19.774 & \light{17.585 - 22.032} \\
            0 & 0 & 17.112 & \light{15.259 - 19.021} & - & - & - & - \\
            \midrule
            1 & 0.055 & 14.354 & \light{12.779 - 16.08} & 33.017 & \light{30.351 - 35.699} & 36.011 & \light{33.027 - 38.934} \\
            2 & 0.087 & 16.804 & \light{15.018 - 18.58} & 36.782 & \light{34.151 - 39.78} & 31.867 & \light{29.142 - 34.636} \\
            3 & 0.115 & 16.946 & \light{15.279 - 18.68} & 35.924 & \light{33.233 - 38.771} & 32.36 & \light{29.821 - 34.87} \\
            4 & 0.137 & 17.598 & \light{15.9 - 19.5} & 32.543 & \light{29.642 - 35.231} & 29.509 & \light{26.957 - 32.077} \\
            5 & 0.157 & 19.35 & \light{17.56 - 21.2} & 25.487 & \light{22.815 - 28.265} & 21.428 & \light{19.138 - 23.791} \\
            10 & 0.221 & 21.538 & \light{19.677 - 23.281} & 11.457 & \light{9.762 - 13.288} & 9.758 & \light{8.234 - 11.331} \\
            20 & 0.297 & 22.414 & \light{20.36 - 24.42} & 10.378 & \light{8.664 - 12.171} & 9.808 & \light{8.146 - 11.343} \\
            100 & 0.482 & 18.195 & \light{16.4 - 20.062} & 3.779 & \light{2.684 - 4.97} & 2.953 & \light{2.103 - 3.853} \\

            \bottomrule
    \end{tabular}
    \label{tab:ndoc_qampari_mistral_bge}
\end{table}

\begin{table}[ht]
    \centering
    \caption{Correctness and citation quality on QAMPARI achieved with Mistral with various numbers of ColBERT retrieved documents, $k$, included in the prompt.}
    \begin{tabular}{rrrrrrrr}
        \toprule
             \multicolumn{1}{c}{} & \multicolumn{1}{c}{\textbf{Ret.}} & \multicolumn{2}{c}{\textbf{EM Recall}} & \multicolumn{2}{c}{\textbf{Citation Recall}} & \multicolumn{2}{c}{\textbf{Citation Precision}} \\
            \cmidrule(r){2-2} 
            \cmidrule(r){3-4}
            \cmidrule(r){5-6}
            \cmidrule(r){7-8}
            k & \multicolumn{1}{c}{Rec@k} & \multicolumn{1}{c}{Mean} & \multicolumn{1}{c}{95\% CI} & \multicolumn{1}{c}{Mean} & \multicolumn{1}{c}{95\% CI} & \multicolumn{1}{c}{Mean} & \multicolumn{1}{c}{95\% CI} \\
            \midrule

            gold & 1 & 39.152 & \light{36.879 - 41.44} & 23.239 & \light{20.827 - 25.708} & 19.774 & \light{17.585 - 22.032} \\
            0 & 0 & 17.112 & \light{15.259 - 19.021} & - & - & - & - \\
            \midrule
            1 & 0.067 & 16.649 & \light{14.88 - 18.461} & 45.446 & \light{42.724 - 48.362} & 50.431 & \light{47.383 - 53.375} \\
            2 & 0.119 & 20.556 & \light{18.72 - 22.44} & 43.893 & \light{41.037 - 46.795} & 36.006 & \light{33.398 - 38.801} \\
            3 & 0.159 & 21.407 & \light{19.48 - 23.3} & 41.728 & \light{38.943 - 44.541} & 36.444 & \light{34.06 - 39.093} \\
            4 & 0.192 & 22.379 & \light{20.54 - 24.3} & 31.825 & \light{29.095 - 34.731} & 27.747 & \light{25.325 - 30.264} \\
            5 & 0.22 & 24.657 & \light{22.5 - 26.64} & 23.981 & \light{21.59 - 26.459} & 20.101 & \light{17.861 - 22.51} \\
            10 & 0.412 & 29.836 & \light{27.68 - 31.962} & 10.084 & \light{8.391 - 11.862} & 8.456 & \light{6.94 - 10.001} \\
            20 & 0.412 & 30.613 & \light{28.359 - 32.9} & 8.417 & \light{6.859 - 10.063} & 6.874 & \light{5.533 - 8.274} \\
            100 & 0.641 & 24.105 & \light{21.94 - 26.381} & 2.381 & \light{1.603 - 3.17} & 1.755 & \light{1.141 - 2.475} \\

            \bottomrule
    \end{tabular}
    \label{tab:ndoc_qampari_mistral_colbert}
\end{table}

\begin{table}[ht]
    \centering
    \caption{Correctness and citation quality on QAMPARI achieved with LLaMA with various numbers of BGE-base retrieved documents, $k$, included in the prompt.}
    \begin{tabular}{rrrrrrrr}
        \toprule
             \multicolumn{1}{c}{} & \multicolumn{1}{c}{\textbf{Ret.}} & \multicolumn{2}{c}{\textbf{EM Recall}} & \multicolumn{2}{c}{\textbf{Citation Recall}} & \multicolumn{2}{c}{\textbf{Citation Precision}} \\
            \cmidrule(r){2-2} 
            \cmidrule(r){3-4}
            \cmidrule(r){5-6}
            \cmidrule(r){7-8}
            k & \multicolumn{1}{c}{Rec@k} & \multicolumn{1}{c}{Mean} & \multicolumn{1}{c}{95\% CI} & \multicolumn{1}{c}{Mean} & \multicolumn{1}{c}{95\% CI} & \multicolumn{1}{c}{Mean} & \multicolumn{1}{c}{95\% CI} \\
            \midrule

           gold & 1 & 32.444 & \light{30.36 - 34.8} & 24.308 & \light{21.621 - 26.906} & 16.42 & \light{14.395 - 18.422} \\
            0 & 0 & 13.725 & \light{12.2 - 15.26} & - & - & - & - \\
            \midrule
            1 & 0.055 & 14.029 & \light{12.58 - 15.56} & 40.375 & \light{37.366 - 43.518} & 38.634 & \light{35.803 - 41.521} \\
            2 & 0.087 & 14.693 & \light{13.14 - 16.32} & 43.257 & \light{40.15 - 46.217} & 31.847 & \light{29.341 - 34.352} \\
            3 & 0.115 & 14.04 & \light{12.62 - 15.502} & 56.025 & \light{53.052 - 59.194} & 38.662 & \light{36.274 - 41.237} \\
            4 & 0.137 & 15.744 & \light{14.22 - 17.28} & 33.229 & \light{30.63 - 35.999} & 22.892 & \light{20.558 - 25.194} \\
            5 & 0.157 & 15.677 & \light{14.02 - 17.3} & 22.242 & \light{19.811 - 24.751} & 16.372 & \light{14.373 - 18.386} \\
            10 & 0.221 & 18.742 & \light{16.88 - 20.62} & 3.443 & \light{2.444 - 4.513} & 2.76 & \light{1.974 - 3.588} \\
            20 & 0.412 & 30.613 & \light{28.359 - 32.9} & 8.417 & \light{6.859 - 10.063} & 6.874 & \light{5.533 - 8.274} \\
            100 & 0.641 & 24.105 & \light{21.94 - 26.381} & 2.381 & \light{1.603 - 3.17} & 1.755 & \light{1.141 - 2.475} \\

            \bottomrule
    \end{tabular}
    \label{tab:ndoc_qampari_llama_bge}
\end{table}

\begin{table}[ht]
    \centering
    \caption{Correctness and citation quality on QAMPARI achieved with LLaMA with various numbers of ColBERT retrieved documents, $k$, included in the prompt.}
    \begin{tabular}{rrrrrrrr}
        \toprule
             \multicolumn{1}{c}{} & \multicolumn{1}{c}{\textbf{Ret.}} & \multicolumn{2}{c}{\textbf{EM Recall}} & \multicolumn{2}{c}{\textbf{Citation Recall}} & \multicolumn{2}{c}{\textbf{Citation Precision}} \\
            \cmidrule(r){2-2} 
            \cmidrule(r){3-4}
            \cmidrule(r){5-6}
            \cmidrule(r){7-8}
            k & \multicolumn{1}{c}{Rec@k} & \multicolumn{1}{c}{Mean} & \multicolumn{1}{c}{95\% CI} & \multicolumn{1}{c}{Mean} & \multicolumn{1}{c}{95\% CI} & \multicolumn{1}{c}{Mean} & \multicolumn{1}{c}{95\% CI} \\
            \midrule

            gold & 1 & 32.444 & \light{30.36 - 34.8} & 24.308 & \light{21.621 - 26.906} & 16.42 & \light{14.395 - 18.422} \\
            0 & 0 & 13.725 & \light{12.2 - 15.26} & - & - & - & - \\
            \midrule
            1 & 0.067 & 16.153 & \light{14.68 - 17.78} & 52.048 & \light{48.95 - 55.218} & 50.288 & \light{47.517 - 53.223} \\
            2 & 0.119 & 18.804 & \light{17.08 - 20.52} & 55.04 & \light{52.239 - 57.998} & 38.711 & \light{35.952 - 41.13} \\
            3 & 0.159 & 19.874 & \light{18.24 - 21.56} & 62.55 & \light{59.748 - 65.626} & 41.091 & \light{38.741 - 43.48} \\
            4 & 0.192 & 21.358 & \light{19.64 - 23.18} & 35.157 & \light{32.209 - 38.061} & 25.063 & \light{22.702 - 27.476} \\
            5 & 0.22 & 21.722 & \light{20.02 - 23.58} & 22.076 & \light{19.738 - 24.713} & 15.817 & \light{13.845 - 17.859} \\
            10 & 0.412 & 24.998 & \light{23.02 - 27.04} & 3.647 & \light{2.533 - 4.746} & 2.723 & \light{1.88 - 3.622} \\

            \bottomrule
    \end{tabular}
    \label{tab:ndoc_qampari_llama_colbert}
\end{table}

\clearpage

\subsection{Additional varied number of gold documents results}
\label{app:addl_gold}

\begin{figure}[ht]
    \begin{center}
        \includegraphics[width=0.7\textwidth]{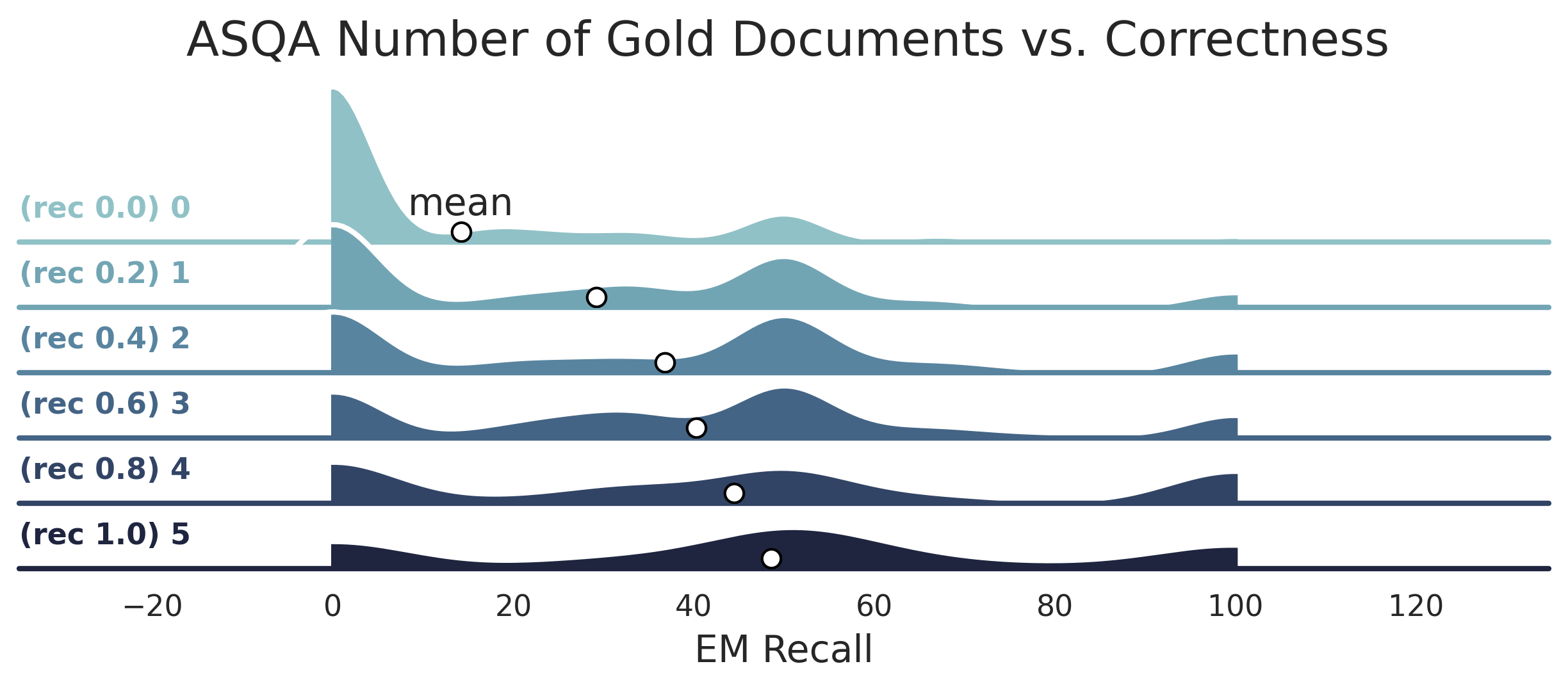}
    \end{center}
    \caption{The per-query relationship between the number of gold documents included in the prompt and the QA accuracy achieved with LLaMA on ASQA. We find that including just one gold document significantly improves accuracy. There is a correlation between the number of gold documents and the accuracy. }
    \label{fig:recall_acc_llama}
\end{figure}

\begin{figure}[ht]
    \begin{center}
        \includegraphics[width=0.7\textwidth]{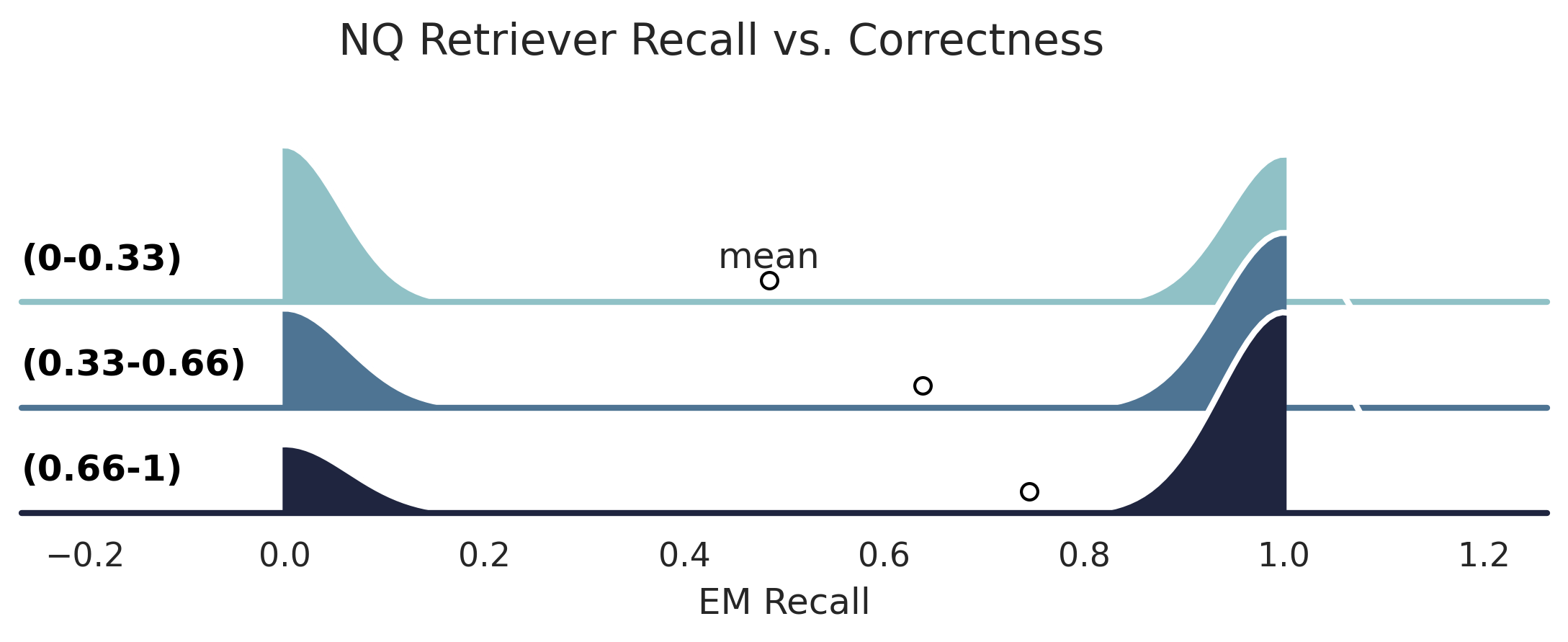}
    \end{center}
    \caption{The per-query relationship between the gold document recall in the prompt and the QA accuracy achieved with Mistral on NQ. There is a correlation between the number of gold documents and the accuracy. }
    \label{fig:recall_acc_nq_mistral}
\end{figure}

Figure \ref{fig:recall_acc_llama} shows the relationship between the number of gold documents in the prompt and correctness achieved on ASQA with LLaMA. In Figure \ref{fig:recall_acc_llama} and \ref{fig:recall_acc_nq_mistral} we present results comparing retriever recall and correctness.

\begin{figure}[ht]
    \begin{center}
        \includegraphics[width=0.7\textwidth]{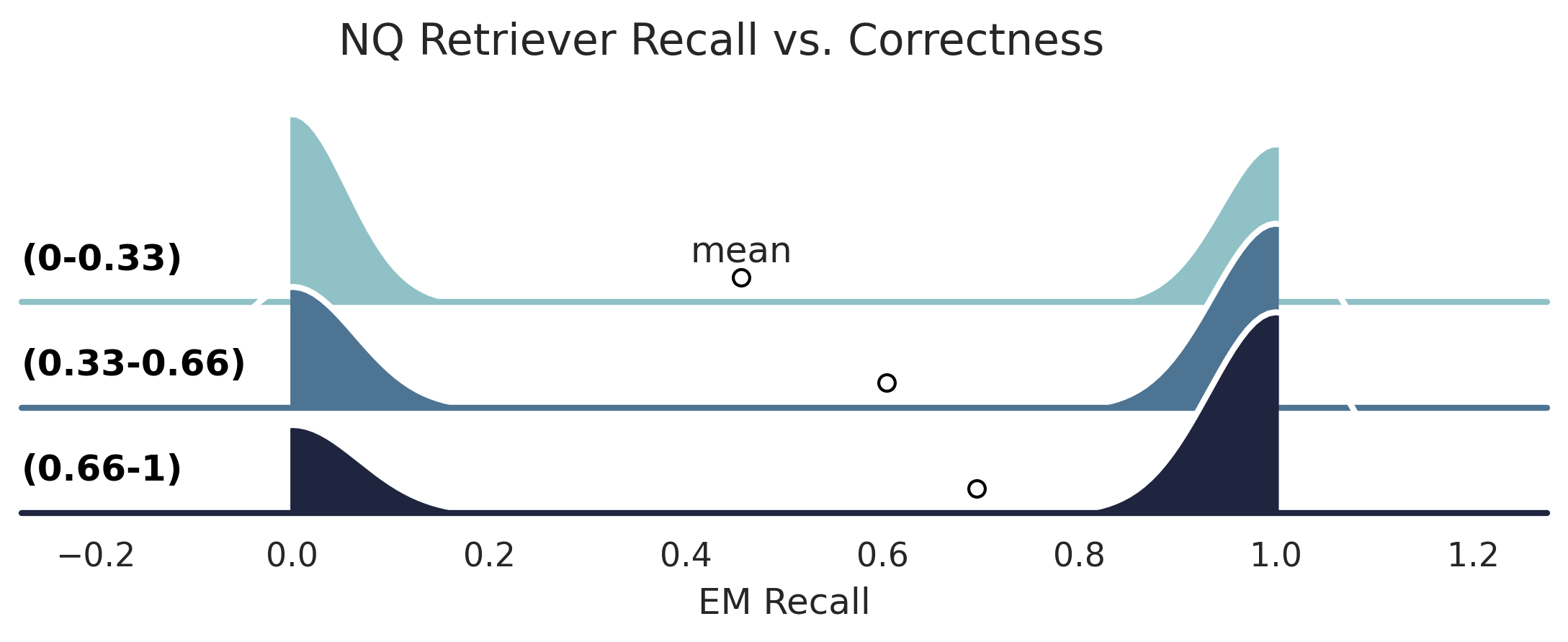}
    \end{center}
    \caption{The per-query relationship between the number of gold documents included in the prompt and the QA accuracy achieved with LLaMA on NQ. We find that including just one gold document significantly improves accuracy. There is a correlation between the number of gold documents and the accuracy. }
    \label{fig:recall_acc_nq_llama}
\end{figure}
\clearpage



\subsection{Additional recall manipulation results}
\label{app:addl_recall}

\begin{figure}[h!]
\centering
    \includegraphics{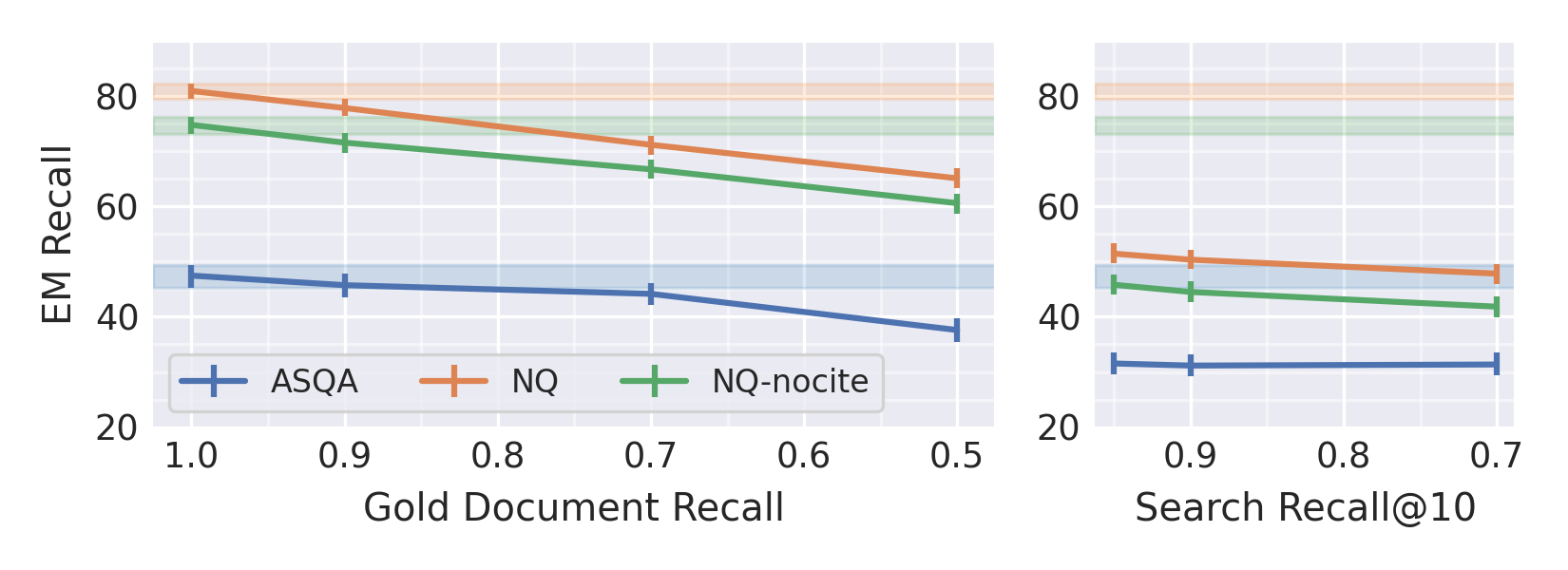}
    \caption{Llama results varying gold document recall (left) and BGE-base search recall (right). Shaded bar is ceiling performance using all gold documents per query. Error bars are 95\% bootstrap confidence intervals.}
    \label{fig:gold_search_recall}
\end{figure}

Generally, both citation recall and citation precision decrease as document recall and search recall decrease.

Since the ASQA dataset is more likely to contain multiple gold evidence documents per query, it is less consistently affected by decreases in document recall. For example, we see in Table \ref{tab:mistral_gold-search-recall_cite} that between 0.7 and 0.9 search recall@10, citation recall is nearly identical for ASQA -- the 95\% CIs are nearly completely overlapping. However, this is not the case for the NQ dataset, which shows a consistent decrease as recall drops.

\begin{table}[ht]
    \centering
    \caption{Full Mistral results for changes in citation metrics as gold document recall and search recall (BGE-base retriever) vary.}
    \resizebox{\textwidth}{!}{
        \begin{tabular}{@{}cllll@{}}
        \toprule
        & \multicolumn{2}{c}{\textbf{Citation Recall}} & \multicolumn{2}{c}{\textbf{Citation Precision}} \\ \midrule
        \multicolumn{1}{l}{} & \multicolumn{1}{c}{ASQA} & \multicolumn{1}{c}{NQ} & \multicolumn{1}{c}{ASQA} & \multicolumn{1}{c}{NQ} \\
        \midrule
        \multicolumn{1}{l}{\textbf{Doc. Recall@10}} &  &  &  &  \\ \midrule
    1.0 & 65.229 (62.876, 67.321) & 75.041 (73.772, 76.397) & 62.375 (60.257, 64.469) & 64.904 (63.660, 66.105) \\
    0.9 & 64.797 (62.701, 66.887) & 72.084 (70.765, 73.409) & 62.092 (60.075, 64.032) & 63.023 (61.723, 64.317) \\
    0.7 & 64.914 (62.781, 66.991) & 66.669 (65.258, 68.148) & 61.599 (59.471, 63.646) & 59.059 (57.693, 60.474) \\
    0.5 & 60.482 (58.238, 62.692) & 60.827 (59.428, 62.292) & 58.731 (56.339, 61.011) & 54.530 (53.122, 55.875) \\
        \midrule
        \multicolumn{1}{l}{\textbf{Search Recall@10}} &  &  &  &  \\ \midrule
        0.95 & 57.636 (55.250, 59.867) & 55.340 (53.872, 56.814) & 52.909 (50.559, 55.129) & 45.906 (44.476, 47.292) \\
        0.9 & 55.764 (53.341, 58.244) & 52.443 (50.835, 53.972) & 50.460 (48.284, 52.665) & 43.263 (42.005, 44.552) \\
        0.7 & 56.017 (53.551, 58.342) & 49.856 (48.405, 51.340) & 52.007 (49.744, 54.265) & 41.523 (40.198, 42.859) \\ \bottomrule
        \end{tabular}
    }
    \label{tab:mistral_gold-search-recall_cite}
\end{table}

\begin{table}[ht]
    \centering
    \caption{Full Llama results for changes in citation metrics as gold document recall and search recall (BGE-base retriever) vary.}
    \resizebox{\textwidth}{!}{
        \begin{tabular}{@{}cllll@{}}
        \toprule
        & \multicolumn{2}{c}{\textbf{Citation Recall}} & \multicolumn{2}{c}{\textbf{Citation Precision}} \\ \midrule
        \multicolumn{1}{l}{} & \multicolumn{1}{c}{ASQA} & \multicolumn{1}{c}{NQ} & \multicolumn{1}{c}{ASQA} & \multicolumn{1}{c}{NQ} \\
        \midrule
        \multicolumn{1}{l}{\textbf{Doc. Recall@10}} &  &  &  &  \\ \midrule
        1.0 & 46.326 (44.123, 48.524) & 54.697 (53.309, 56.104) & 46.294 (44.082, 48.686) & 55.748 (54.285, 57.262) \\
        0.9 & 46.778 (44.492, 48.999) & 53.438 (51.979, 54.862) & 46.411 (44.130, 48.779) & 55.758 (54.275, 57.143) \\
        0.7 & 42.450 (40.073, 44.566) & 49.078 (47.618, 50.660) & 42.618 (40.377, 45.055) & 50.265 (48.662, 51.771) \\
        0.5 & 39.634 (37.468, 41.970) & 45.827 (44.258, 47.346) & 39.780 (37.453, 41.911) & 47.506 (45.929, 48.962) \\
        \midrule
        \multicolumn{1}{l}{\textbf{Search Recall@10}} &  &  &  &  \\ \midrule
        0.95 & 34.425 (32.106, 36.668) & 28.412 (27.054, 29.697) & 30.218 (28.282, 32.411) & 25.735 (24.561, 26.914) \\
        0.9 & 33.419 (31.234, 35.618) & 26.626 (25.383, 27.994) & 28.745 (26.698, 30.792) & 24.402 (23.262, 25.549) \\
        0.7 & 34.190 (31.658, 36.568) & 25.033 (23.710, 26.321) & 29.509 (27.349, 31.602) & 22.853 (21.708, 24.054) \\ \bottomrule
        \end{tabular}
    }
    \label{tab:llama_gold-search-recall_cite}
\end{table}

\subsection{Additional noise experiment results}
\label{app:addl_noise}

In Table \ref{tab:noise_percentile_asqa_mistral} we show detailed results for ASQA Mistral performance after injecting noisy documents from various percentiles of similarity to the query. These correspond to the correctness results in Figure \ref{fig:noise_percentile_asqa_mistral} in the main body of the paper. 

Table \ref{tab:first100_noise_asqa_nl} shows the first 100 noise experiment for ASQA with Llama2 for augmenting both gold and BGE-base retrieved data with noise first in the prompt.

\begin{table}[ht]
    \centering
    \caption{ASQA Mistral performance with gold or BGE-base retrieved documents (respectively) \textit{and} noisy documents from various percentiles of similarity to the query. We find that adding noisy documents from all percentiles degrades both correctness and citation performance. There is no obvious correlation between the percentile of the noise and the degradation of performance.}
    \subfloat[5 Gold docs]{
        \begin{tabular}{rrrr}
            \toprule
            \multicolumn{1}{l}{} & \multicolumn{1}{c}{\textbf{Correct.}} & \multicolumn{2}{c}{\textbf{Citation}} \\
            \cmidrule(r){2-2}
            \cmidrule(r){3-4}
            \multicolumn{1}{l}{noise percentile} & \multicolumn{1}{l}{EM Rec.} & \multicolumn{1}{l}{Rec.} & \multicolumn{1}{l}{Prec.} \\
            \midrule
            -   & 50.673 & 65.403 & 62.462 \\
            10  & 40.500 & 53.000 & 55.298 \\
            20  & 42.533 & 55.243 & 59.944 \\
            30  & 42.983 & 56.767 & 55.459 \\
            40  & 43.483 & 53.833 & 56.917 \\
            50  & 42.633 & 53.783 & 55.800 \\
            60  & 39.733 & 53.157 & 54.985 \\
            70  & 42.433 & 57.750 & 57.429 \\
            80  & 42.833 & 56.283 & 56.075 \\
            90  & 43.517 & 59.533 & 62.917 \\
            100 & 42.333 & 57.417 & 60.245 \\
            \bottomrule
        \end{tabular}%
        
        }
    \quad
    \subfloat[5 docs retrieved with BGE-base]{
            \begin{tabular}{rrr}
                \toprule
                 \multicolumn{1}{c}{\textbf{Correct.}} & \multicolumn{2}{c}{\textbf{Citation}}\\
                 \cmidrule(r){1-1}
                \cmidrule(r){2-3}
                \multicolumn{1}{l}{EM Rec.} & \multicolumn{1}{l}{Rec.} & \multicolumn{1}{l}{Prec.} \\
                \midrule
                37.569 & 61.188 & 58.763 \\
                34.550 & 50.286 & 52.583 \\
                34.683 & 52.967 & 53.254 \\
                34.100 & 51.133 & 49.812 \\
                32.783 & 50.350 & 58.433 \\
                35.300 & 54.250 & 54.867 \\
                31.783 & 50.036 & 51.439 \\
                29.833 & 53.667 & 55.573 \\
                32.467 & 50.900 & 51.321 \\
                32.000 & 49.900 & 49.900 \\
                34.250 & 50.900 & 52.275 \\
                \bottomrule
            \end{tabular}%
        
        }
    \label{tab:noise_percentile_asqa_mistral}
\end{table}

\begin{table}[ht]
    \caption{LLaMA performance on ASQA when adding non-gold (\textit{noise}) documents based on their similarity ranking (between $5^{\text{th}}-100^{\text{th}}$ nearest neighbor). BGE-base results (right) with 5 retrieved documents. Noisy documents are added after the gold or retrieved documents in the prompt.}
    \centering
    \resizebox{\textwidth}{!}{
    \subfloat[5 gold docs]{
        \label{asqa_nl_first100_noise_gold}
        \begin{tabular}{lrrr}
        \toprule
         & \multicolumn{1}{c}{\textbf{Correct.}} & \multicolumn{2}{c}{\textbf{Citation}}  \\
        \cmidrule(r){2-2}
        \cmidrule(r){3-4}
        noise idx. & \multicolumn{1}{l}{EM Rec.} & \multicolumn{1}{l}{Rec.} & \multicolumn{1}{l}{Prec.} \\
        \midrule
        gold only & 47.426 & 46.261 & 46.033 \\
        gold + $5^{\text{th}}-10^{\text{th}}$ & 43.070 & 40.778 & 36.135 \\
        gold + $95^{\text{th}}-100^{\text{th}}$ & 41.807 & 38.646 & 35.536 \\
        \bottomrule
        \end{tabular}%
        }
    \quad
    \subfloat[5 BGE-base retrieved docs]{
        \label{asqa_nl_first100_noise_bge}
        \begin{tabular}{lrrr}
        \toprule
        & \multicolumn{1}{c}{\textbf{Correct.}} & \multicolumn{2}{c}{\textbf{Citation}}  \\
        \cmidrule(r){2-2}
        \cmidrule(r){3-4}
        noise idx. & \multicolumn{1}{l}{EM Rec.} & \multicolumn{1}{l}{Rec.} & \multicolumn{1}{l}{Prec.} \\
        \midrule
        BGE-base only & 34.488 & 42.029 & 41.781 \\
        BGE-base + $5^{\text{th}}-10^{\text{th}}$  & 33.284 & 33.979 & 30.298 \\
        BGE-base + $95^{\text{th}}-100^{\text{th}}$ & 30.979 & 34.561 & 31.932 \\
        \bottomrule
        \end{tabular}%
        }
    }
    \label{tab:first100_noise_asqa_nl}
\end{table}

%% file: 0_rx-main.bbl
\begin{thebibliography}{10}

\bibitem{Lewis_2020}
Patrick Lewis, Ethan Perez, Aleksandra Piktus, Fabio Petroni, Vladimir Karpukhin, Naman Goyal, Heinrich K\"{u}ttler, Mike Lewis, Wen-tau Yih, Tim Rockt\"{a}schel, Sebastian Riedel, and Douwe Kiela.
\newblock Retrieval-augmented generation for knowledge-intensive nlp tasks.
\newblock In {\em Proceedings of the 34th International Conference on Neural Information Processing Systems}, NIPS '20, Red Hook, NY, USA, 2020. Curran Associates Inc.

\bibitem{huang2023surveyhallucinationlargelanguage}
Lei Huang, Weijiang Yu, Weitao Ma, Weihong Zhong, Zhangyin Feng, Haotian Wang, Qianglong Chen, Weihua Peng, Xiaocheng Feng, Bing Qin, and Ting Liu.
\newblock A survey on hallucination in large language models: Principles, taxonomy, challenges, and open questions, 2023.

\bibitem{wang2024knowledgeeditinglargelanguage}
Song Wang, Yaochen Zhu, Haochen Liu, Zaiyi Zheng, Chen Chen, and Jundong Li.
\newblock Knowledge editing for large language models: A survey, 2024.

\bibitem{Rashkin_Nikolaev_Lamm_Aroyo_Collins_Das_Petrov_Tomar_Turc_Reitter_2023}
Hannah Rashkin, Vitaly Nikolaev, Matthew Lamm, Lora Aroyo, Michael Collins, Dipanjan Das, Slav Petrov, Gaurav~Singh Tomar, Iulia Turc, and David Reitter.
\newblock Measuring attribution in natural language generation models.
\newblock {\em Computational Linguistics}, 49(4):777–840, December 2023.

\bibitem{Gao_Yen_Yu_Chen_2023}
Tianyu Gao, Howard Yen, Jiatong Yu, and Danqi Chen.
\newblock Enabling large language models to generate text with citations.
\newblock (arXiv:2305.14627), October 2023.
\newblock arXiv:2305.14627 [cs].

\bibitem{liu-etal-2023-evaluating}
Nelson Liu, Tianyi Zhang, and Percy Liang.
\newblock Evaluating verifiability in generative search engines.
\newblock In Houda Bouamor, Juan Pino, and Kalika Bali, editors, {\em Findings of the Association for Computational Linguistics: EMNLP 2023}, pages 7001--7025, Singapore, December 2023. Association for Computational Linguistics.

\bibitem{Cuconasu_2024}
Florin Cuconasu, Giovanni Trappolini, Federico Siciliano, Simone Filice, Cesare Campagnano, Yoelle Maarek, Nicola Tonellotto, and Fabrizio Silvestri.
\newblock The power of noise: Redefining retrieval for {RAG} systems.
\newblock In {\em Proceedings of the 47th International ACM SIGIR Conference on Research and Development in Information Retrieval}, SIGIR 2024. ACM, July 2024.

\bibitem{yu2024evaluationretrievalaugmentedgenerationsurvey}
Hao Yu, Aoran Gan, Kai Zhang, Shiwei Tong, Qi~Liu, and Zhaofeng Liu.
\newblock Evaluation of retrieval-augmented generation: A survey, 2024.

\bibitem{meng2023locatingeditingfactualassociations}
Kevin Meng, David Bau, Alex Andonian, and Yonatan Belinkov.
\newblock Locating and editing factual associations in gpt, 2023.

\bibitem{dale2022detectingmitigatinghallucinationsmachine}
David Dale, Elena Voita, Loïc Barrault, and Marta~R. Costa-jussà.
\newblock Detecting and mitigating hallucinations in machine translation: Model internal workings alone do well, sentence similarity even better, 2022.

\bibitem{guoSemanticModelsFirststage2022}
Jiafeng Guo, Yinqiong Cai, Yixing Fan, Fei Sun, Ruqing Zhang, and Xueqi Cheng.
\newblock Semantic {{Models}} for the {{First-stage Retrieval}}: {{A Comprehensive Review}}.
\newblock 40(4):1--42.

\bibitem{zhaoDenseTextRetrieval2024}
Wayne~Xin Zhao, Jing Liu, Ruiyang Ren, and Ji-Rong Wen.
\newblock Dense {{Text Retrieval Based}} on {{Pretrained Language Models}}: {{A Survey}}.
\newblock 42(4):89:1--89:60.

\bibitem{santhanamColBERTv2EffectiveEfficient2022}
Keshav Santhanam, Omar Khattab, Jon Saad-Falcon, Christopher Potts, and Matei Zaharia.
\newblock {{ColBERTv2}}: {{Effective}} and {{Efficient Retrieval}} via {{Lightweight Late Interaction}}.
\newblock In {\em Proceedings of the 2022 {{Conference}} of the {{North American Chapter}} of the {{Association}} for {{Computational Linguistics}}: {{Human Language Technologies}}}. arXiv.

\bibitem{muennighoff2022mteb}
Niklas Muennighoff, Nouamane Tazi, Lo{\"\i}c Magne, and Nils Reimers.
\newblock Mteb: Massive text embedding benchmark.
\newblock {\em arXiv preprint arXiv:2210.07316}, 2022.

\bibitem{thakur2021beir}
Nandan Thakur, Nils Reimers, Andreas R{\"u}ckl{\'e}, Abhishek Srivastava, and Iryna Gurevych.
\newblock {BEIR}: A heterogeneous benchmark for zero-shot evaluation of information retrieval models.
\newblock In {\em Thirty-fifth Conference on Neural Information Processing Systems Datasets and Benchmarks Track (Round 2)}, 2021.

\bibitem{Yoran_Wolfson_Ram_Berant_2024}
Ori Yoran, Tomer Wolfson, Ori Ram, and Jonathan Berant.
\newblock Making retrieval-augmented language models robust to irrelevant context.
\newblock (arXiv:2310.01558), May 2024.
\newblock arXiv:2310.01558 [cs].

\bibitem{tepper2023leanvec}
Mariano Tepper, Ishwar~Singh Bhati, Cecilia Aguerrebere, Mark Hildebrand, and Ted Willke.
\newblock {LeanVec}: Search your vectors faster by making them fit.
\newblock {\em Transactions onf Machine Learning Research}, 2024.

\bibitem{touvron2023llamaopenefficientfoundation}
Hugo Touvron, Thibaut Lavril, Gautier Izacard, Xavier Martinet, Marie-Anne Lachaux, Timothée Lacroix, Baptiste Rozière, Naman Goyal, Eric Hambro, Faisal Azhar, Aurelien Rodriguez, Armand Joulin, Edouard Grave, and Guillaume Lample.
\newblock Llama: Open and efficient foundation language models, 2023.

\bibitem{jiang2023mistral7b}
Albert~Q. Jiang, Alexandre Sablayrolles, Arthur Mensch, Chris Bamford, Devendra~Singh Chaplot, Diego de~las Casas, Florian Bressand, Gianna Lengyel, Guillaume Lample, Lucile Saulnier, Lélio~Renard Lavaud, Marie-Anne Lachaux, Pierre Stock, Teven~Le Scao, Thibaut Lavril, Thomas Wang, Timothée Lacroix, and William~El Sayed.
\newblock Mistral 7b, 2023.

\bibitem{bge_embedding}
Shitao Xiao, Zheng Liu, Peitian Zhang, and Niklas Muennighoff.
\newblock C-pack: Packaged resources to advance general chinese embedding, 2023.

\bibitem{aguerrebere_similarity_2023}
Cecilia Aguerrebere, Ishwar~Singh Bhati, Mark Hildebrand, Mariano Tepper, and Theodore Willke.
\newblock Similarity {Search} in the {Blink} of an {Eye} with {Compressed} {Indices}.
\newblock {\em Proc. VLDB Endow.}, 16(11):3433--3446, July 2023.

\bibitem{DBLP:journals/corr/abs-2004-12832}
Omar Khattab and Matei Zaharia.
\newblock Colbert: Efficient and effective passage search via contextualized late interaction over {BERT}.
\newblock {\em CoRR}, abs/2004.12832, 2020.

\bibitem{stelmakh-etal-2022-asqa}
Ivan Stelmakh, Yi~Luan, Bhuwan Dhingra, and Ming-Wei Chang.
\newblock {ASQA}: Factoid questions meet long-form answers.
\newblock In Yoav Goldberg, Zornitsa Kozareva, and Yue Zhang, editors, {\em Proceedings of the 2022 Conference on Empirical Methods in Natural Language Processing}, pages 8273--8288, Abu Dhabi, United Arab Emirates, December 2022. Association for Computational Linguistics.

\bibitem{amouyal-etal-2023-qampari}
Samuel Amouyal, Tomer Wolfson, Ohad Rubin, Ori Yoran, Jonathan Herzig, and Jonathan Berant.
\newblock {QAMPARI}: A benchmark for open-domain questions with many answers.
\newblock In Sebastian Gehrmann, Alex Wang, Jo{\~a}o Sedoc, Elizabeth Clark, Kaustubh Dhole, Khyathi~Raghavi Chandu, Enrico Santus, and Hooman Sedghamiz, editors, {\em Proceedings of the Third Workshop on Natural Language Generation, Evaluation, and Metrics (GEM)}, pages 97--110, Singapore, December 2023. Association for Computational Linguistics.

\bibitem{kwiatkowski-etal-2019-natural}
Tom Kwiatkowski, Jennimaria Palomaki, Olivia Redfield, Michael Collins, Ankur Parikh, Chris Alberti, Danielle Epstein, Illia Polosukhin, Jacob Devlin, Kenton Lee, Kristina Toutanova, Llion Jones, Matthew Kelcey, Ming-Wei Chang, Andrew~M. Dai, Jakob Uszkoreit, Quoc Le, and Slav Petrov.
\newblock Natural questions: A benchmark for question answering research.
\newblock {\em Transactions of the Association for Computational Linguistics}, 7:452--466, 2019.

\bibitem{Hsia_Shaikh_Wang_Neubig_2024}
Jennifer Hsia, Afreen Shaikh, Zhiruo Wang, and Graham Neubig.
\newblock Ragged: Towards informed design of retrieval augmented generation systems.
\newblock (arXiv:2403.09040), March 2024.
\newblock arXiv:2403.09040 [cs].

\bibitem{petroni2021kiltbenchmarkknowledgeintensive}
Fabio Petroni, Aleksandra Piktus, Angela Fan, Patrick Lewis, Majid Yazdani, Nicola~De Cao, James Thorne, Yacine Jernite, Vladimir Karpukhin, Jean Maillard, Vassilis Plachouras, Tim Rocktäschel, and Sebastian Riedel.
\newblock Kilt: a benchmark for knowledge intensive language tasks, 2021.

\bibitem{hesterbergBootstrap2011}
Tim Hesterberg.
\newblock Bootstrap.
\newblock 3(6):497--526, 2011.

\bibitem{BehnamGhader_Miret_Reddy_2023}
Parishad BehnamGhader, Santiago Miret, and Siva Reddy.
\newblock Can retriever-augmented language models reason? the blame game between the retriever and the language model.
\newblock In {\em Findings of the Association for Computational Linguistics: EMNLP 2023}, page 15492–15509, Singapore, 2023. Association for Computational Linguistics.

\bibitem{chenBenchmarkingLargeLanguage2024}
Jiawei Chen, Hongyu Lin, Xianpei Han, and Le~Sun.
\newblock Benchmarking {{Large Language Models}} in {{Retrieval-Augmented Generation}}.
\newblock 38(16):17754--17762.

\bibitem{izacardLeveragingPassageRetrieval2021}
Gautier Izacard and Edouard Grave.
\newblock Leveraging {{Passage Retrieval}} with {{Generative Models}} for {{Open Domain Question Answering}}.
\newblock In Paola Merlo, Jorg Tiedemann, and Reut Tsarfaty, editors, {\em Proceedings of the 16th {{Conference}} of the {{European Chapter}} of the {{Association}} for {{Computational Linguistics}}: {{Main Volume}}}, pages 874--880. Association for Computational Linguistics.

\bibitem{Salemi_Zamani_2024}
Alireza Salemi and Hamed Zamani.
\newblock Evaluating retrieval quality in retrieval-augmented generation.
\newblock (arXiv:2404.13781), April 2024.
\newblock arXiv:2404.13781 [cs].

\bibitem{li_approximate_2020}
Wen Li, Ying Zhang, Yifang Sun, Wei Wang, Mingjie Li, Wenjie Zhang, and Xuemin Lin.
\newblock Approximate {Nearest} {Neighbor} {Search} on {High} {Dimensional} {Data} — {Experiments}, {Analyses}, and {Improvement}.
\newblock {\em IEEE Transactions on Knowledge and Data Engineering}, 32(8):1475--1488, August 2020.
\newblock Conference Name: IEEE Transactions on Knowledge and Data Engineering.

\bibitem{malkov_efficient_2020}
Yu~A. Malkov and D.~A. Yashunin.
\newblock Efficient and {Robust} {Approximate} {Nearest} {Neighbor} {Search} {Using} {Hierarchical} {Navigable} {Small} {World} {Graphs}.
\newblock {\em IEEE Transactions on Pattern Analysis and Machine Intelligence}, 42(4):824--836, 2020.

\bibitem{subramanya_diskann_2019}
Suhas~Jayaram Subramanya, Devvrit, Rohan Kadekodi, Ravishankar Krishnaswamy, and Harsha Simhadri.
\newblock {DiskANN}: {Fast} {Accurate} {Billion}-point {Nearest} {Neighbor} {Search} on a {Single} {Node}.
\newblock In {\em Advances in {Neural} {Information} {Processing} {Systems}}, 2019.

\end{thebibliography}
